\documentclass[10pt,twocolumn,letterpaper]{article}

\newif \ifarxiv
\arxivtrue

\newif \ifmainpaper
\mainpapertrue

\ifarxiv
\usepackage[pagenumbers]{cvpr} 
\else
\usepackage{cvpr}              
\fi

\usepackage{graphicx}
\usepackage{amsmath}
\usepackage{amssymb}
\usepackage{booktabs}
\usepackage{times}
\usepackage{algorithm2e} 
\usepackage{enumitem} 
\usepackage{euscript} 
\usepackage{pifont} 
\usepackage{multirow} 
\usepackage[accsupp]{axessibility}  

\usepackage[pagebackref,breaklinks,colorlinks]{hyperref}

\usepackage[capitalize]{cleveref}
\crefname{section}{Sec.}{Secs.}
\Crefname{section}{Section}{Sections}
\Crefname{table}{Table}{Tables}
\crefname{table}{Tab.}{Tabs.}

\def\cvprPaperID{5003} 
\def\confName{CVPR}
\def\confYear{2023}

\makeatletter
\newcommand{\printfnsymbol}[1]{%
  \textsuperscript{\@fnsymbol{#1}}%
}
\makeatother

\makeatletter
\renewcommand\@makefntext[1]{\leftskip=0em\hskip-0em\@makefnmark#1}
\makeatother

\long\def\ignorethis#1{}

\usepackage{color}
\definecolor{gray}{rgb}{0.6,0.6,0.6}
\definecolor{red}{rgb}{1,0,0}
\definecolor{green}{rgb}{0,1,0}
\definecolor{blue}{rgb}{0,0,1}
\definecolor{dark-green}{rgb}{0,0.4,0}
\definecolor{orange}{rgb}{1,0.55,0}
\definecolor{white}{rgb}{1,1,1}
\definecolor{black}{rgb}{0,0,0}
\definecolor{dark-brown}{rgb}{0.2,0.1,0}
\definecolor{light-blue}{rgb}{0.4,0.6,0.99}
\definecolor{dark-red}{rgb}{0.6,0,0}
\definecolor{light-red}{rgb}{1,0.2,0.6}
\definecolor{pink}{rgb}{1,0.2,0.6}
\definecolor{dark-pink}{rgb}{0.6,0,0.3}

\newcommand{\whitetxt}[1]{{\color{white}#1}\normalfont}

\newcommand{\camrdy}[1]{{\color{black}#1}\normalfont}

\newbox\jsavebox

\newcommand{\Lcal}{\mathcal{L}}
\newcommand{\Dcal}{\mathcal{D}}
\newcommand{\Neu}{\EuScript{N}}

\newcommand{\CorrFlow}{SCOOP }
\newcommand{\CorrFlowwo}{SCOOP}

\newcommand{\cmark}{\text{\ding{51}}}
\newcommand{\xmark}{\text{\ding{55}}}

\DeclareMathOperator*{\argmin}{\arg\!\min}
\DeclareMathOperator{\diag}{diag}

\begin{document}

\title{\CorrFlowwo: Self-Supervised Correspondence and Optimization-Based Scene Flow}

\author{
Itai Lang$^{1,2}$\printfnsymbol{1} \quad \quad Dror Aiger$^{2}$ \quad \quad Forrester Cole$^{2}$ \quad \quad Shai Avidan$^{1}$ \quad \quad Michael Rubinstein$^{2}$ \\
$^{1}$Tel Aviv University \quad \quad $^{2}$Google Research \\
{\tt\small \{itailang@mail, avidan@eng\}.tau.ac.il ~ \quad \quad \{aigerd, fcole, mrub\}@google.com} ~~
}

\maketitle

\begin{abstract}
Scene flow estimation is a long-standing problem in computer vision, where the goal is to find the 3D motion of a scene from its consecutive observations. Recently, there have been efforts to compute the scene flow from 3D point clouds. A common approach is to train a regression model that consumes source and target point clouds and outputs the per-point translation vector. An alternative is to learn point matches between the point clouds concurrently with regressing a refinement of the initial correspondence flow. In both cases, the learning task is very challenging since the flow regression is done in the free 3D space, and a typical solution is to resort to a large annotated synthetic dataset.

We introduce \CorrFlowwo, a new method for scene flow estimation that can be learned on a small amount of data without employing ground-truth flow supervision. In contrast to previous work, we train a pure correspondence model focused on learning point feature representation and initialize the flow as the difference between a source point and its softly corresponding target point. Then, in the run-time phase, we directly optimize a flow refinement component with a self-supervised objective, which leads to a coherent and accurate flow field between the point clouds. Experiments on widespread datasets demonstrate the performance gains achieved by our method compared to existing leading techniques while using a fraction of the training data. Our code is publicly available\footnote{\url{https://github.com/itailang/SCOOP} \\ \printfnsymbol{1}The work was done during an internship at Google Research.}.
\end{abstract}

\section{Introduction} \label{sec:introduction}

Scene flow estimation~\cite{vedula1999three} is a fundamental problem in computer vision with various use-cases, such as autonomous driving, scene parsing, pose estimation, and object tracking, to name a few. Given two consecutive observations of a 3D scene, the aim is to compute the dynamics of the scene between the observations. Scene flow prediction based on 2D images has been thoroughly investigated in the literature~\cite{wedel2008efficient, wedel2011stereoscopic, vogel2013piecewise, menze2015object, ma2019deep}. However, in light of the recent proliferation of 3D sensors, such as LiDAR, there is a surge of interest in scene flow methods that operate directly on the 3D data~\cite{liu2019flownet3d, gu2019hplflownet, mittal2020just, wu2020pointpwcnet, li2021self}.

\begin{figure}[t!]
\centering
\includegraphics[width=\linewidth]{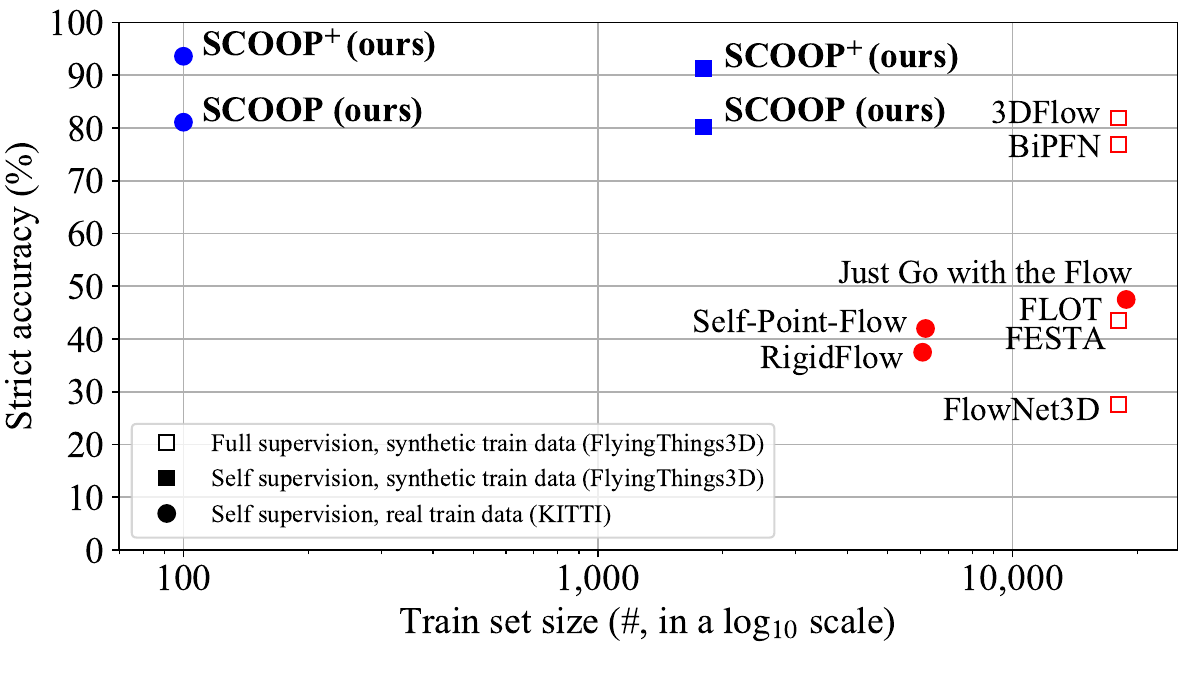}
\vspace{-20pt}
\caption{\textbf{Flow accuracy on the KITTI benchmark \vs the train set size.} \camrdy{Our method} is trained on one or two orders of magnitude less data while surpassing the performance of the competing \camrdy{techniques}~\cite{liu2019flownet3d, mittal2020just, puy2020flot, wang2021festa, li2021self, li2022rigidflow, wang2022what, cheng2022bi-pointflownet} by a large margin. Please see Table~\ref{tbl:kitti} for the complete details of the evaluation settings.}
\label{fig:teaser_scoop}
\end{figure}

Liu \etal~\cite{liu2019flownet3d} were among the first to pursue this research avenue. They proposed FlowNet3D, a fully-supervised neural network that learned to regress the flow between 3D point clouds and showed remarkable performance improvement over image-based techniques~\cite{brox2011large, vogel20153d, menze2015object}. Since their method required ground-truth flow annotations, which are scarce for real-world data, they turned to training on a large synthetic dataset that compromised the generalization capability to real-world LiDAR data.

Follow-up works devised self-supervised learning schemes~\cite{mittal2020just, li2021self} and narrowed the domain gap by training on unannotated LiDAR point cloud pairs. However, similar to Liu \etal~\cite{liu2019flownet3d}, they used a regression approach in which the model should learn to compute the flow in the \textit{free 3D space}. This task is extremely challenging, given the irregular nature of point clouds, and requires a large amount of training data for the network to converge.

\begin{figure*}[tb!]
\centering
\begin{tabular}{c c c c}
\multicolumn{4}{c}{\includegraphics[width=0.97\linewidth]{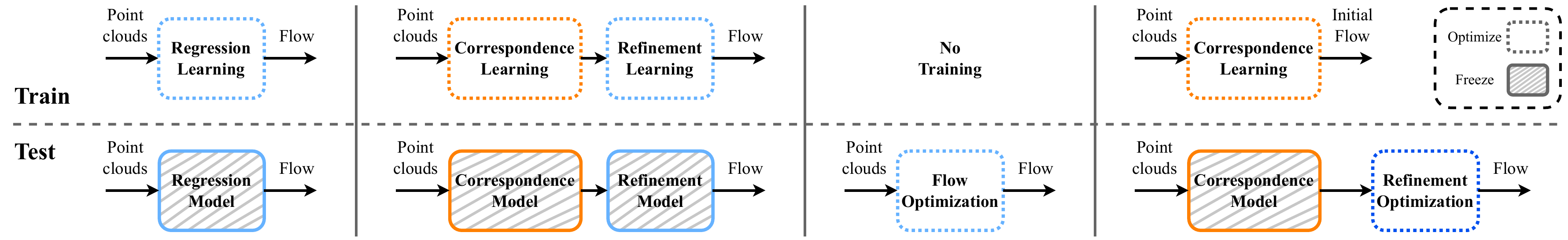}} \\
\whitetxt{aaaaaaa}FlowNet3D~\cite{liu2019flownet3d} & \whitetxt{aaaaaaaaaa}FLOT~\cite{puy2020flot} & 
\whitetxt{aaaaaaaaaaa}Neural Prior~\cite{li2021neural} & \whitetxt{a}\CorrFlow (ours) \\
\end{tabular}
\vspace{-5pt}
\caption{\textbf{Comparison of scene flow approaches.} Given a pair of point clouds, FlowNet3D~\cite{liu2019flownet3d} learns to regress the flow in the free 3D space, and the trained model is frozen for testing. FLOT~\cite{puy2020flot} concurrently trains two network components: one that computes point correspondence and another that regresses a correction to the resulting correspondence flow. Neural Prior~\cite{li2021neural} optimizes the flow between the point clouds from scratch without learning. In contrast to previous work, we take a hybrid approach. We train a \textit{pure correspondence model without flow regression}, which serves for flow initialization. Then, we \textit{directly optimize only the flow refinement} at the test-time.}
\label{fig:approach_comparison}
\end{figure*}

In another line of work~\cite{puy2020flot, kittenplon2021flowstep3d, gojcic2021weakly}, researchers leveraged point cloud correspondence for scene flow prediction. In this approach, the flow is computed as the translation of a point in the first point cloud (source) to its softly corresponding point in the second one (target). The softly corresponding point is a weighted sum of target points based on point similarity in a learned latent space. Thus, rather than the challenging regression problem in the 3D ambient space, the flow estimation task boils down to point feature learning and is reduced to the convex combination space~\cite{rockafellar1970convex} of existing target points. However, to relax this constraint, another network component is trained to regress flow corrections. The joint \camrdy{training} of point representation and flow refinement burdens the \camrdy{learning} process and retains the reliance on large datasets with flow supervision.

Another emerging approach is an optimization-only flow computation~\cite{pontes2020scene, li2021neural}. In this case, no training data is involved, and the flow is optimized at run-time for each scene separately. Despite the high accuracy such a dedicated optimization achieves, it requires a long processing time.

We present \CorrFlowwo, a hybrid flow estimation method that can be learned from a small amount of training data. \CorrFlow consists of two parts: a self-supervised neural network for point cloud correspondence and a direct flow refinement optimization module. During the training phase, the network learns to extract point features for soft point matches, which initialize the flow between the point clouds. In contrast to previous work, our network is focused on learning just the point embeddings, allowing its training on a very small dataset, as shown in Figure~\ref{fig:teaser_scoop}. Additionally, we consider the confidence of the network in the computed correspondences to guide the learning process better.

Then, instead of training another network for regressing flow updates, we define an optimization problem and directly optimize residual flow refinement vectors at run-time. The optimization objective encourages a coherent flow field while retaining the translated source points close to the target point cloud. Our design choices improve the accuracy compared to learning-based methods and reduce the processing time with respect to the optimization-only approach~\cite{pontes2020scene, li2021neural}. For both correspondence learning and refinement optimization, we use a self-supervised distance objective and a smoothness prior instead of ground-truth flow labels. Figure~\ref{fig:approach_comparison} presents the difference between our approach and leading previous ones.

In summary, we propose a hybrid flow prediction approach for point clouds based on self-supervised correspondence learning and direct run-time residual flow optimization. Using well-established datasets in the scene flow literature, we show that our approach yields clear performance improvement over existing state-of-the-art methods while using a fraction of the training data and without employing any ground-truth flow supervision.

\section{Related Work} \label{sec:related_work}

\noindent \textbf{Flow regression.} \quad A common approach for scene flow estimation on point clouds is to train a flow regression model~\cite{liu2019flownet3d, gu2019hplflownet, wu2020pointpwcnet, wang2021festa, cheng2022bi-pointflownet}. It is a neural network that computes the flow vectors between the point clouds in the ambient 3D space. Liu \etal~\cite{liu2019flownet3d} proposed FlowNet3D, which encoded the point clouds into a latent space, mixed point features with a flow embedding layer, and regressed the scene flow by decoding the mixed point features. FlowNet3D was trained in a fully-supervised manner, using an $l_2$ loss with respect to ground-truth flow annotations.

Liu \etal~\cite{liu2019flownet3d} inspired a line of follow-up works~\cite{mittal2020just, wang2021festa, li2021self, wang2022what, cheng2022bi-pointflownet}. Wang \etal~\cite{wang2021festa} added spatial and temporal attention layers to FlowNet3D's architecture. In Bi-PointFlowNet~\cite{cheng2022bi-pointflownet}, the authors propagated features from each point cloud bidirectionally, augmenting the point feature representation. Mittal \etal~\cite{mittal2020just} discarded flow supervision by utilizing a self-supervised nearest neighbor loss and cycle consistency between the forward and reveres scene flows, and Li \etal~\cite{li2021self, li2022rigidflow} extracted flow labels for training from the data itself. Similar to the latter methods, we also refrain from ground-truth flow supervision in our training scheme. However, rather than flow regression, we base our technique on soft point matches in the scene, which simplifies the flow estimation problem.

\medskip
\noindent \textbf{Point cloud correspondence.} \quad
\camrdy{Finding correspondences is widely applied to various vision tasks~\cite{lang2021dpc, puy2020flot, wu2022sc-wls, zhong2022snake}.} Several methods \camrdy{have been} proposed for dense mapping between non-rigid \camrdy{point cloud} shapes~\cite{groueix20183dcoded, deprelle2019learning, zeng2021corrnet3d, lang2021dpc}. Recently, Lang \etal~\cite{lang2021dpc} suggested constructing one point cloud by the other using latent space similarity and the point coordinates themselves rather than regressing the corresponding point cloud~\cite{groueix20183dcoded, zeng2021corrnet3d}. Inspired by Lang's work, we do not use flow regression in our model and concentrate the learning process on point feature representation. However, while Lang \etal operated on complete shapes with one-to-one correspondence, our method accommodates scenes with partial objects where a perfect match may not exist.

Researchers have taken the correspondence approach to the scene flow problem as well~\cite{puy2020flot, kittenplon2021flowstep3d, gojcic2021weakly}. FLOT~\cite{puy2020flot} computed an optimal transport plan that served \camrdy{for} an initial flow between the point clouds and further regressed flow refinement with a series of learned convolutions. Our work builds on FLOT but differs from it in three main aspects. First, we exclude flow regression from our training scheme and instead apply direct run-time optimization to refine the initial correspondence-based flow. Second, we use the model's confidence in the computed point matches to improve the point feature learning. Third, we do not use any ground-truth flow annotations, neither for the correspondence training nor for the refinement optimization, whereas FLOT relies on fully-supervised scene flow data.

\medskip
\noindent \textbf{Optimization-based scene flow.} \quad Pontes \etal~\cite{pontes2020scene} suggested a scene flow estimation technique that does not involve learning. Instead, the flow was optimized completely at run-time, such that the warped source is close to the target point cloud while demanding the flow to be ``as-rigid-as-possible". Pontes \etal encoded this prior by minimizing the graph Laplacian defined over the source points. In follow-up work~\cite{li2021neural}, the explicit graph was replaced by a neural prior, which implicitly regularized the optimized flow field. In contrast to these papers, we initialize the flow with a learned correspondence model and optimize only the residual flow refinement at run-time.

\section{Method} \label{sec:method}
A point cloud is a set of unordered 3D points $X \in \mathbb{R}^{n \times 3}$, where $n$ is the number of points. Given a pair of point clouds of a scene, denoted as $X, Y \in \mathbb{R}^{n \times 3}$ and referred to as source and target, respectively, our goal is to estimate a flow field $F^* \in \mathbb{R}^{n \times 3}$ describing the per-point motion from $X$ to $Y$.

\begin{figure}[tb!]
\centering
\includegraphics[width=\linewidth]{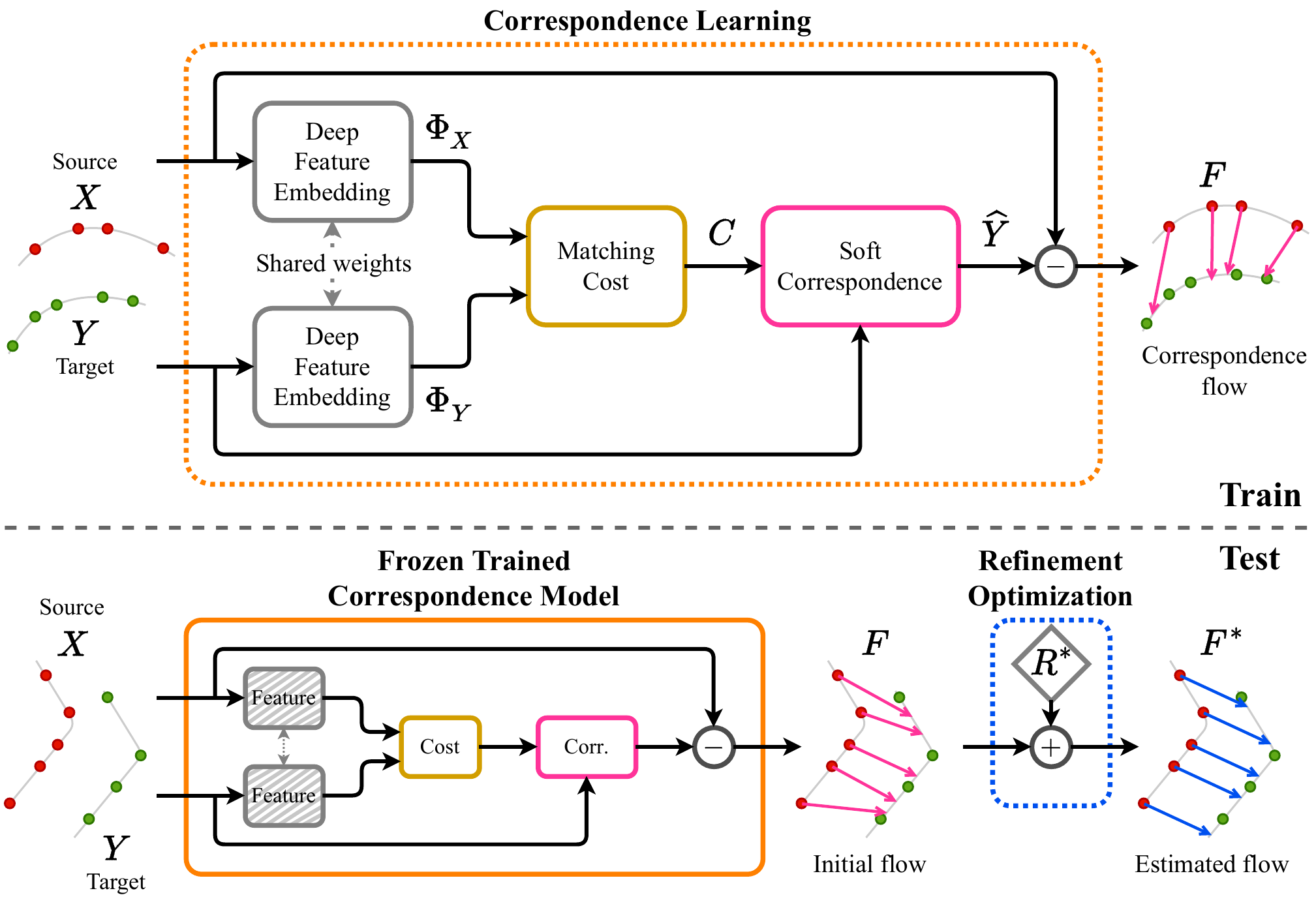}
\caption{\textbf{The proposed method.} \CorrFlow includes two components: a learned point cloud correspondence model and a flow refinement module. The model learns deep point embeddings $\Phi_X, \Phi_Y$ to establish soft point matches based on a matching cost $C$ in the latent space. The initial flow $F$ from the training phase is the difference between the softly corresponding point cloud $\widehat{Y}$ and the source point cloud $X$. At \camrdy{the} test-time, we freeze the trained model and optimize a residual flow refinement $R^*$ to produce a smooth and consistent scene flow $F^*$ between the point clouds.}
\label{fig:system}
\end{figure}

We tackle this problem via self-supervised soft correspondence learning between the two point clouds and a direct flow refinement optimization. An overview of the method is shown in Figure~\ref{fig:system}. First, a deep neural network is used to extract point features. Then, we calculate a matching cost between points in the learned feature space. Based on this cost, we solve an optimal transport problem to compute a softly matched target point for each source point, where the difference between the two is regarded as the correspondence-based flow. Finally, we refine the flow field by demanding its consistency across neighboring source points and obtain our estimated scene flow. In both correspondence learning and flow refinement, no ground-truth flow labels are \camrdy{employed}.

\subsection{Matching Cost} \label{subsec:matching_cost}
The cost of matching a point $x_i \in X$ to a point $y_j \in Y$ is determined based on the point representation learned by a deep neural network. The network consumes the raw point clouds $X$, $Y$ and computes point features $\Phi_X, \Phi_Y \in \mathbb{R}^{n \times d}$, where $d$ is the per-point feature dimension. The network's architecture is based on PointNet++~\cite{qi2017pointnetpp}. Its details are given in the supplemental material.

Inspired by previous work~\cite{puy2020flot, li2021self, lang2021dpc}, we first compute the cosine similarity in the learned feature space:
\begin{equation} \label{eq:similarity_scoop}
S_{ij} = \frac{\Phi_X^i \cdot (\Phi_Y^j)^\top}{||\Phi_X^i||_2 ||\Phi_Y^j||_2},
\end{equation}

\noindent where $\Phi_X^i, \Phi_Y^j \in \mathbb{R}^{d}$ are the $i$'th and $j$'th rows of $\Phi_X$ and $\Phi_Y$, respectively. Then, the cost is set to
\begin{equation} \label{eq:cost}
C_{ij} = 1 - S_{ij}
\end{equation}
\noindent for points with a Euclidean distance less than 10 meters and to $\infty$ otherwise to avoid flow between points too far apart.

\subsection{Soft Correspondence} \label{subsec:soft_corr}
Finding correspondence between the source and target point clouds can be modeled as an optimal transport problem, where each source point is assigned with a mass $\frac{1}{n}$ that is transported to the target points~\cite{puy2020flot, li2021self}. \camrdy{Similar to} FLOT~\cite{puy2020flot}, we use the relaxed transport problem:
\begin{equation} \label{eq:regularized_transport}
\begin{split}
T^* = & \argmin_{T \in \mathbb{R}_{+}^{n \times n}} \sum_{ij} (C_{ij} T_{ij} + \epsilon T_{ij} (\log T_{ij} - 1)) \\
& + \lambda (\mathsf{KL}(T 1_n, \frac{1}{n} 1_n) + \mathsf{KL}(T^\top 1_n, \frac{1}{n} 1_n)),
\end{split}
\end{equation}

\noindent where $C_{ij} \ge 0$ is the matching cost from Equation~\ref{eq:cost} and $T_{ij} \ge 0$ is the amount of mass transported between points. The parameters $\epsilon, \lambda \ge 0$ control the relaxation of the problem. $1_n \in \mathbb{R}^n$ is a vector with all entries equal 1. $\mathsf{KL}$ is the Kullback-Leibler divergence used for soft preservation of the transported mass between the point clouds.

The second term in the summation operation in Equation~\ref{eq:regularized_transport} is an entropic regularization, which enables solving the problem efficiently by the Sinkhorn algorithm~\cite{cuturi2013sinkhorn, chizat2018scaling}. We use this algorithm to estimate the optimal transport matrix $T^*$ from $C$ to represent the soft correspondence between the point clouds. The complete derivation of the transport problem and the Sinkhorn algorithm's details are given in the supplementary material.

\subsection{Correspondence-Based Flow} \label{subsec:point_correspondence}
We leverage the optimal transport plan $T^*$ to compute correspondence weights for the source and target points for an initial estimate of the scene flow. Different from FLOT~\cite{puy2020flot}, which includes all the target points as candidates for each source point, we consider only target points with maximal transport amount from the source point. This design choice focuses our flow estimation pipeline on the most relevant target candidates and improves the method's results.

For a point $x_i \in X$, the matching weights are calculated as follows:
\begin{equation} \label{eq:corr_weights}
w_{ij} = \frac{e^{T^*_{ij}}}{\sum_{l \in \Neu_Y (x_i)} e^{T^*_{il}}},
\end{equation}

\noindent where $\Neu_Y(x_i)$ is a neighborhood containing the $k_s$ indices of the $\{y_j\}$ points with the top mass transport $\{T^*_{ij}\}$. The softly corresponding point $\hat{y}_{x_i}$ to $x_i$ is:
\begin{equation} \label{eq:soft_corr}
\hat{y}_{x_i} = \sum_{j \in \Neu_Y (x_i)}{w_{ij} y_j},
\end{equation}

\noindent and the \camrdy{initial} estimated flow for the point $x_i$ is:
\begin{equation} \label{eq:corr_flow}
f_i  = \hat{y}_{x_i} - x_i.
\end{equation}

\noindent Note that if we define $\widehat{T}^*_{ij} = w_{ij}$ for $j \in \Neu_Y (x_i)$ and $0$ otherwise, we get the initial flow field as:
\begin{equation} \label{eq:flow_field}
F = \widehat{T}^* Y - X = \widehat{Y} - X,
\end{equation}

\noindent where $\widehat{Y} \in \mathbb{R}^{n \times 3}$ contains the points $\{\hat{y}_{x_i}\}$.

\subsection{Training Objective} \label{subsec:training_objective}
To learn point representation suitable for scene flow without ground-truth supervision, we apply the flowing loss terms. First, for a tractable flow estimation, we would like each softly corresponding point $\hat{y}_{x_i}$ to have a nearby target point $y_j$. It may be achieved by the nearest-neighbor distance term, as done by Mittal \etal~\cite{mittal2020just}:
\begin{equation} \label{eq:nn_loss}
\Dcal = \frac{1}{|X|}\sum_{x_i \in X}{\min_{y_j \in Y}||\hat{y}_{x_i} - y_j||_2^2}.
\end{equation}

\noindent However, the correspondence quality for the source points can vary. For example, points on a flat region will have less distinctive correspondences than points with geometrically unique features. Thus, we augment the distance term in Equation~\ref{eq:nn_loss} with the matching confidence of each point.

The confidence measure is based on the correspondence similarity that we define as:
\begin{equation} \label{eq:corr_conf}
s_{x_i} = \sum_{j \in \Neu_Y (x_i)}{w_{ij} S_{ij}}.
\end{equation}

\noindent The value of $s_{x_i}$ is in the range $[-1, 1]$. To get a confidence value between $0$ and $1$, we trim the negative values, set the matching confidence of $x_i$ to be $p_{x_i} = \max(s_{x_i}, 0)$, and use $p_{x_i}$ to define our confidence-aware distance loss:
\begin{equation} \label{eq:dist_loss}
\Lcal_{dist} = \frac{1}{|X|}\sum_{x_i \in X} p_{x_i} {\min_{y_j \in Y} ||\hat{y}_{x_i} - y_j||_2^2}.
\end{equation}

The loss term $\Lcal_{dist}$ can be minimized by either minimizing $p_{x_i}$ or the distance between $\hat{y}_{x_i}$ and its nearest neighbor $y_j \in Y$. To avoid the degenerate solution of $p_{x_i} = 0$ for all $x_i \in X$, we add a confidence loss term:
\begin{equation} \label{eq:conf_loss}
\Lcal_{conf} = \frac{1}{|X|}\sum_{x_i \in X} 1 - p_{x_i},
\end{equation}

\noindent which penalizes the degenerate solution.

Additionally, to preserve the geometric structure of the source point cloud, we would like the flow field to be smooth. That is, neighboring source points should have a similar flow prediction. Thus, we regularize the learning process with a flow smoothness loss~\cite{kittenplon2021flowstep3d}:
\begin{equation} \label{eq:flow_loss}
\Lcal_{flow} = \frac{1}{|X|k_f}\sum_{x_i \in X}\sum_{l \in N_X(x_i)} {||f_i - f_l||_1},
\end{equation}

\noindent where $N_X(x_i)$ is the Euclidean neighborhood of $x_i$ in $X \setminus x_i$ of size $k_f$. The overall training objective is:
\begin{equation} \label{eq:total_loss_scoop}
\Lcal_{total} = \Lcal_{dist} + \alpha_{conf} \Lcal_{conf} + \alpha_{flow} \Lcal_{flow},
\end{equation}

\noindent where $\alpha_{conf}$ and $\alpha_{flow}$ are hyperparameters, balancing the contribution of the different loss terms.

\subsection{Flow Refinement Optimization} \label{subsec:flow_refinement}
The advantage of the correspondence-based flow, presented in Equation~\ref{eq:flow_field}, is that the softly matching points are in the vicinity of the surface of objects in the target scene. However, it limits the flow to the convex hull~\cite{rockafellar1970convex} of points in the target point cloud. We enable the flow to deviate from this constraint by a flow refinement optimization step at run-time.

\begin{figure}[t!]
\centering
\includegraphics[width=\linewidth]{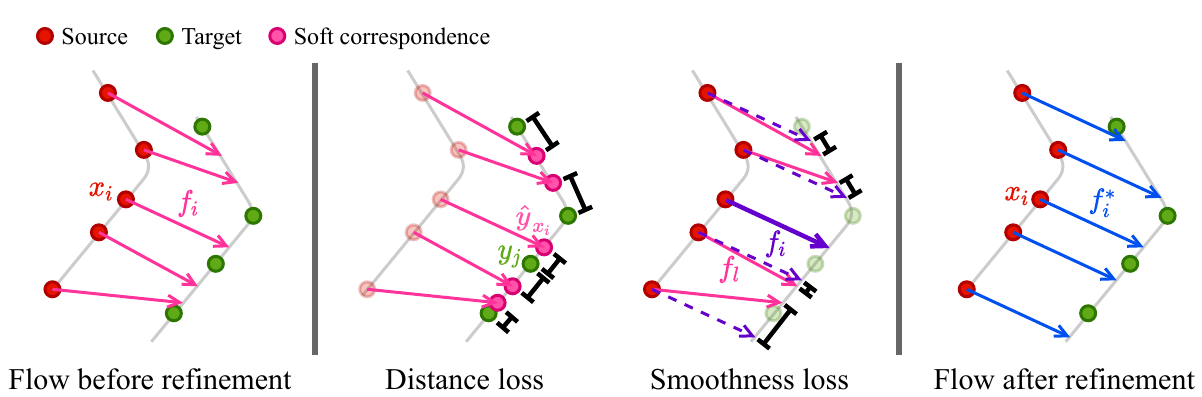}
\caption{\textbf{Illustration of the flow refinement objective.} The initial flow $\{f_i\}$ stems from the translation of the source points $\{x_i\}$ (red) to their softly corresponding ones $\{\hat{y}_{x_i}\}$ (magenta). The flow is refined with a distance loss that keeps the proximity of the translated points to the target points $\{y_j\}$ (green) and a smoothness loss that encourages similar flow vectors (dashed purple) for neighboring points. The optimization process results in a flow field $\{f_i^*\}$  (blue) that preserves the structure of the source point cloud and warps it close to the implicit surface of the target point cloud.}
\label{fig:losses}
\end{figure}

Instead of training an additional neural network part to regress flow corrections, as done by Puy \etal~\cite{puy2020flot}, we \textit{directly} optimize a flow refinement component $R^* \in \mathbb{R}^{n \times 3}$ using the self-supervised distance and smoothness losses defined in Equations~\ref{eq:dist_loss} and \ref{eq:flow_loss}, respectively. An illustration of these losses is depicted in Figure~\ref{fig:losses}.

The optimization problem for the flow refinement takes the form:
\begin{equation} \label{eq:flow_refinement}
\begin{split}
R^* = \argmin_{R \in \mathbb{R}^{n \times 3}}\frac{1}{|X|}&\sum_{x_i \in X}{\min_{y_j \in Y}p_{x_i}||x_i + (f_i + r_i) - y_j||_2^2} \\
+ \lambda_{flow}\frac{1}{|X|k_f}&\sum_{x_i \in X}\sum_{l \in N_X(x_i)} {||(f_i + r_i) - (f_l + r_l)||_1},
\end{split}
\end{equation}

\noindent where $r_i \in R$ is the flow refinement for point $x_i$, and the refined scene flow is $F^* = F + R^*$. Our flow refinement module further preserves the structure of the source point cloud, where the target points $\{y_j\}$ are used as anchors to guide the refined flow and keep the proximity to the underlying target surface.

\section{Experiments} \label{sec:experiments}

\begin{table*}[t!]
\small
\centering
\begin{tabular}{@{ } l l l c @{ } c @{ } c @{ } c @{ } c @{ }}
\toprule
Method & Supervision & Train data & Test data\whitetxt{a} & $\:EPE\!\downarrow$ & $\:AS\!\uparrow$ & $\:AR\!\uparrow$ & $\:Out.\!\downarrow$ \\
\midrule
FlowNet3D~\cite{liu2019flownet3d} & \textit{Full} & FT3D\textsubscript{o} (18,000) & KITTI\textsubscript{o} & 0.173 & 27.6 & 60.9 & 64.9\\
FLOT~\cite{puy2020flot}           & \textit{Full} & FT3D\textsubscript{o} (18,000) & KITTI\textsubscript{o} & 0.107  & 45.1 & 74.0 & 46.3 \\
FESTA~\cite{wang2021festa}        & \textit{Full} & FT3D\textsubscript{o} (18,000) & KITTI\textsubscript{o} & 0.094 & 44.9 & 83.4 & - \\
3DFlow~\cite{wang2022what} & \textit{Full} & FT3D\textsubscript{o} (18,000) & KITTI\textsubscript{o} & 0.073 & 81.9 & 89.0 & 26.1 \\
BiPFN~\cite{cheng2022bi-pointflownet} & \textit{Full} & FT3D\textsubscript{o} (18,000) & KITTI\textsubscript{o} & 0.065 & 76.9 & 90.6 & 26.4 \\
\camrdy{\CorrFlow (ours)}                              & \camrdy{\textit{Self}} & \camrdy{FT3D\textsubscript{o} (1,800)} & \camrdy{KITTI\textsubscript{o}} &
\camrdy{0.063} & \camrdy{79.7} & \camrdy{91.0} & \camrdy{24.4} \\
\CorrFlowwo\textsuperscript{+} (ours)                              & \textit{Self} & FT3D\textsubscript{o} (1,800) & KITTI\textsubscript{o} &
\textbf{0.047} & \textbf{91.3} & \textbf{95.0} & \textbf{18.6} \\
\midrule
JGF~\cite{mittal2020just} & \textit{Full} + \textit{Self} + \textit{Self} & FT3D\textsubscript{o} (18,000) + nuScenes (700) + KITTI\textsubscript{v} (100) & KITTI\textsubscript{t} & 0.105 & 46.5 & 79.4 & - \\
SPF~\cite{li2021self}     & \textit{Self} + \textit{Self} & KITTI\textsubscript{r} (6,068) + KITTI\textsubscript{v} (100) & KITTI\textsubscript{t} & 0.089 & 41.7 & 75.0 & - \\
RigidFlow~\cite{li2022rigidflow} & \textit{Self} & KITTI\textsubscript{r} (6,068) & KITTI\textsubscript{t} & 0.117 & 38.8 & 69.7 & - \\
\camrdy{\CorrFlow (ours)} & \camrdy{\textit{Self}} & \camrdy{KITTI\textsubscript{v} (100)} & \camrdy{KITTI\textsubscript{t}} &
\camrdy{\textbf{0.052}} & \camrdy{\textbf{80.6}} & \camrdy{\textbf{92.9}} & \camrdy{\textbf{19.7}} \\
\midrule
Graph Prior~\cite{pontes2020scene} & \textit{Self} & N/A (optimization-only) & KITTI\textsubscript{t} & 0.082 & 84.0 & 88.5 & - \\
Neural Prior~\cite{li2021neural} & \textit{Self} & N/A (optimization-only) & KITTI\textsubscript{t} & \textbf{0.036} & 92.3 & 96.2 & - \\
\CorrFlowwo\textsuperscript{+} (ours) & \textit{Self} & KITTI\textsubscript{v} (100) & KITTI\textsubscript{t} &
0.039 & \textbf{93.6} & \textbf{96.5} & \textbf{15.2} \\
\bottomrule
\end{tabular}
\caption{\textbf{Quantitative comparison.} We compare scene flow evaluation metrics for different supervision settings, train data, and test data. The number of training examples is indicated in parentheses. $EPE$, $AS$, $AR$, and $Out.$ stand for End-Point-Error, Strict Accuracy, Relaxed Accuracy, and Outliers, respectively. \camrdy{The symbol \textsuperscript{+} indicates an evaluation using all the points in the test point clouds, as done for the optimization-only methods~\cite{pontes2020scene, li2021neural}.} While other baselines apply fully-supervised training, our method yields better performance without employing ground-truth flow labels. Besides, \CorrFlow can be trained \textit{only} on KITTI\textsubscript{v}, with as few as 100 training instances. In contrast, alternative learning-based methods use additional training data, such as nuScenes, or a large dataset, such as KITTI\textsubscript{r}. \camrdy{Please see further details in subsections~\ref{subsec:experimental_setup} and~\ref{subsec:flow_res}.}}
\label{tbl:kitti}
\end{table*}

In this section, we evaluate \CorrFlowwo's performance using widely spread datasets and compare it with recent state-of-the-art (SOTA) works on scene flow estimation. Additionally, we demonstrate the influence of the flow refinement module, analyze the performance and run-time duration, and verify our design choices with an ablation study.

\subsection{Experimental Setup} \label{subsec:experimental_setup}

\noindent \textbf{Datasets.} \quad We adopt two common datasets in the scene flow literature, FlyingThings3D~\cite{mayer2016large} and KITTI~\cite{menze2015object, menze2015joint}. Originally, these benchmarks did not include point cloud data. They were processed to \camrdy{a} point cloud format by Liu \etal~\cite{liu2019flownet3d} and denoted as FT3D\textsubscript{o} and KITTI\textsubscript{o}, respectively.

FT3D\textsubscript{o} is a large-scale synthetic dataset with 18,000/2,000 train/validation scene examples of randomly moving objects from the ShapeNet collection~\cite{chang2015shapenet}. Each example contains a pair of point clouds and ground-truth flow vectors. Since the objects' motion is randomized, they may appear or disappear from the view of the scene and create occlusions. The dataset also includes a mask for points whose flow is invalid due to occlusions.

The KITTI\textsubscript{o} dataset contains 150 real-world LiDAR scenes. Every scene includes source and target point clouds with flow annotations for the source points. Ground points are removed, and the source points are considered to have a valid flow~\cite{liu2019flownet3d}. KITTI\textsubscript{o} was further split by Mittal \etal~\cite{mittal2020just} into sets of 100 and 50 examples, marked as KITTI\textsubscript{v} and KITTI\textsubscript{t}, respectively, for fine-tuning experiments. Li \etal~\cite{li2021self} also built a large unlabeled LiDAR dataset for self-supervised learning on real-world data. They took raw LiDAR scans from the KITTI scenes~\cite{menze2015object, menze2015joint}, disjoint from the KITTI\textsubscript{o} data, and created a training set of 6,068 instances denoted as KITTI\textsubscript{r}.

\medskip
\noindent \textbf{Evaluation metrics.} \quad We use well-established evaluation metrics from previous works~\cite{liu2019flownet3d, mittal2020just, puy2020flot}: End-Point-Error $EPE\:[m]$, Strict Accuracy $AS\:[\%]$, Relaxed Accuracy $AR\:[\%]$, and Outliers $Out.\:[\%]$. These metrics are based on the point error $e_i$ and the relative error $e_i^{rel}$:
\begin{equation} \label{eq:err}
e_i = || f_i^* - f_i^{gt} ||_2, \quad e_i^{rel} = \frac{|| f_i^* - f_i^{gt} ||_2}{||f_i^{gt} ||_2},
\end{equation}

\noindent where $f_i^*$ and $f_i^{gt}$ are the predicted and ground-truth flow for point $x_i$, respectively. The $EPE$ is the average point error, measured in meters; $AS$ is the \camrdy{percentage} of points whose $e_i < 0.05\:[m]$ or $e_i^{rel} < 5\%$; $AR$ is the \camrdy{percentage} of points for which $e_i < 0.1\:[m]$ or $e_i^{rel} < 10\%$; and $Out.$ is the \camrdy{percentage} of points with $e_i > 0.3\:[m]$ or $e_i^{rel} > 10\%$.

\medskip
\noindent \textbf{Implementation details.} \quad \CorrFlow is implemented in PyTorch~\cite{paszke2017automatic}, where the publicly available PointNet++~\cite{qi2017pointnetpp} implementation is adapted for our point feature embedding. The model is trained on $n = 2,\!048$ points, sampled at random from the source and target point clouds of the scene examples. Only the 3D coordinates of the points are used as input to the model. The parameters $\epsilon$ and $\lambda$ from Equation~\ref{eq:regularized_transport} are defined as learnable variables and optimized as part of the learning process. The point feature dimension is $d=128$. For the neighborhood sizes we use $k_s = 64$, $k_f = 32$, and the losses' hyperparameters are set to $\alpha_{conf} = 0.1$, $\alpha_{flow} = 10$.

\camrdy{As in previous work~\cite{liu2019flownet3d, puy2020flot, mittal2020just, li2022rigidflow}, we evaluate \CorrFlow on point clouds of 2,048 points randomly sampled from the source and target. However, the full point clouds of KITTI\textsubscript{o} and KITTI\textsubscript{t} are an order of magnitude larger and have different cardinality. Thus, for a complete evaluation of the entire scene flow, we also utilize our method (denoted as \CorrFlowwo\textsuperscript{+} for this case) to exploit the whole point cloud information and test the performance for the original resolution.} Additional \camrdy{implementation} details appear in the supplementary.

\medskip
\noindent \textbf{Baseline methods.} \quad Our method is contrasted with the recent methods FlowNet3D~\cite{liu2019flownet3d}, FLOT~\cite{puy2020flot}, FESTA~\cite{wang2021festa}, 3DFlow~\cite{wang2022what}, and BiPFN~\cite{cheng2022bi-pointflownet}. These methods require ground-truth flow supervision. Additionally, we compare our results with the recent self-supervised flow models of Mittal \etal~\cite{mittal2020just} and Li \etal~\cite{li2021self, li2022rigidflow}, and the optimization-based techniques Graph Prior~\cite{pontes2020scene} and Neural Prior~\cite{li2021neural}.

\subsection{Scene Flow Results} \label{subsec:flow_res}

\noindent \textbf{Cross-dataset evaluation.} \quad We demonstrate the generalization power of \CorrFlow by training it on the FT3D\textsubscript{o} and testing its performance on KITTI\textsubscript{o}. Table~\ref{tbl:kitti} summarises the results. The alternative methods~\cite{liu2019flownet3d, puy2020flot, wang2021festa, wang2022what, cheng2022bi-pointflownet} are trained on FT3D\textsubscript{o} in a fully-supervised \camrdy{fashion}: their models are learned with the ground-truth flow information, and the points with an occluded flow are excluded from the training objective using the mask provided in the dataset.

\begin{figure*}[tb!]
\centering
\begin{tabular}{c c}
\multicolumn{2}{c}{\includegraphics[width=0.96\linewidth]{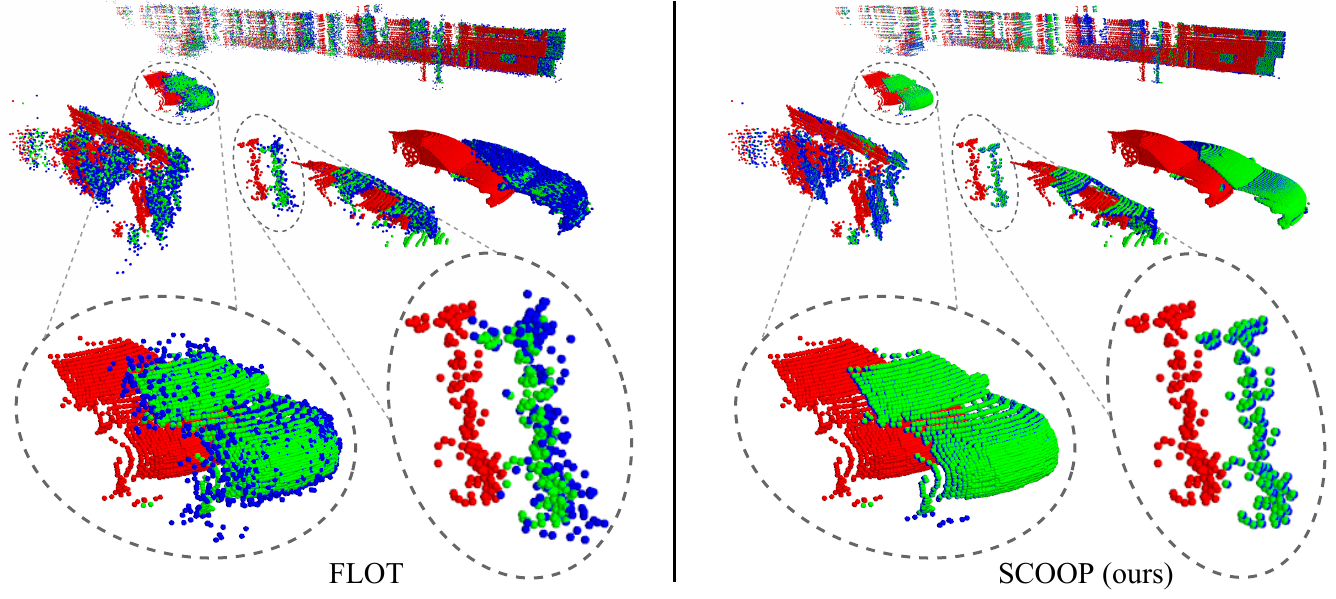}} \\ 
\end{tabular}
\vspace{-10pt}
\caption{\textbf{Visual comparison of scene flow results for a KITTI\textsubscript{o} example scene.} The training was done on the FT3D\textsubscript{o} dataset. The source and target point clouds are shown in red and green, respectively. The warped source point cloud by FLOT~\cite{puy2020flot} (left) and by our method (right) is presented in blue. While the result of FLOT deviates from the surface of the target, \CorrFlow preserves the source point cloud's structure and computes its accurate flow.}
\label{fig:flow_comp}
\end{figure*}

In contrast, our model is trained in a \textit{completely} self-supervised manner. We assume no knowledge of the flow annotations nor the occlusion mask and do not use them in our losses. Additionally, we use \textit{only} 1,800 randomly selected examples from FT3D\textsubscript{o}, while the competitors \camrdy{employ} all 18,000 scene instances. \camrdy{Still, \CorrFlow improves over the SOTA method BiPFN~\cite{cheng2022bi-pointflownet} in all the evaluation metrics. Moreover, utilizing the entire point cloud data further increases our performance.}

FlowNet3D~\cite{liu2019flownet3d}, FESTA~\cite{wang2021festa}, 3DFlow~\cite{wang2022what}, and BiPFN~\cite{cheng2022bi-pointflownet} are regression-based networks that predict the flow in the 3D ambient space. The models adapt to the characteristics of the synthetic training set, and the generalization to the real-world test data is limited. FLOT~\cite{puy2020flot} leverages point cloud correspondence based on learned point features, which eases the flow prediction problem. However, it also jointly learns to regress a flow correction component that burdens the point representation training process.

\CorrFlowwo, on the other hand, is focused only on learning point embeddings suitable for scene flow estimation, guided by our self-supervised losses. It \camrdy{extracts} discriminative features, which transfer well across the FT3D\textsubscript{o} and KITTI\textsubscript{o} datasets, and enables to compute the correspondence-based flow between the point clouds. In contrast to FLOT, we delegate the flow refinement process to the test phase, directly optimize it in a self-supervised fashion, and surpass their flow estimation performance.

Figure~\ref{fig:flow_comp} shows a visual comparison between the results of FLOT and our method. The warped source point cloud by FLOT is noisy, and the structure of objects in the scene is compromised. On the contrary, \CorrFlow produces a coherent flow field across neighboring points, preserves their local geometry, and accurately predicts the scene flow. Additional visualizations are presented in the supplementary.

\medskip
\noindent \textbf{Training on a small dataset.} \quad Since our model does not include a flow regression component and has to learn only point features, it can be trained on a very limited amount of data. To demonstrate this ability, we train it \textit{from scratch} on the 100 point cloud pairs of KITTI\textsubscript{v} and use KITTI\textsubscript{t} for testing. The results of this experiment are presented in Table~\ref{tbl:kitti}.

Different from our work, the competing methods of Mittal \etal~\cite{mittal2020just} and Li \etal~\cite{li2021self, li2022rigidflow} are based on flow regression and require a large amount of training data. Mittal \etal utilize a fully-supervised pre-training on FT3D\textsubscript{o}, and the additional outdoor flow dataset nuScenes~\cite{caesar2020nuscenes}, before fine-tuning on KITTI\textsubscript{v}. Li \etal~\cite{li2021self, li2022rigidflow} train their model on the large KITTI\textsubscript{r} \camrdy{dataset}. Our \CorrFlow outperforms these other methods while being trained \textit{only} on KITTI\textsubscript{v}, which is almost two orders of magnitudes smaller than KITTI\textsubscript{r}.

The pure optimization methods~\cite{pontes2020scene, li2021neural} find the solution per scene separately, which might lead to sub-optimal local minima. In contrast, we leverage the correspondence statistics learned from the data and adapt the initial flow to the scene at hand by our residual run-time optimization. The initial correspondence flow serves as a good starting point for the optimization phase, yielding a \camrdy{similar or} better final result compared to the optimization-only alternatives.

\begin{figure}[t!]
\centering
\includegraphics[width=\linewidth]{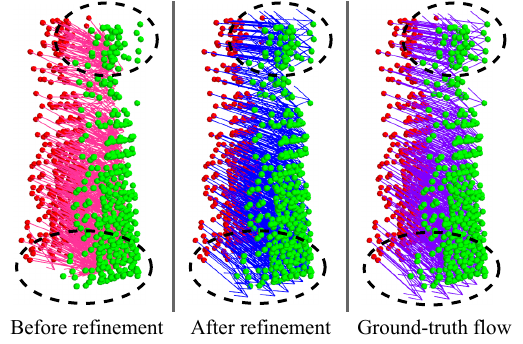}
\vspace{-15pt}
\caption{\textbf{The flow refinement effect.} We demonstrate the effect on data from KITTI\textsubscript{t}. The source point cloud is in red, and the target is in green. Our correspondence model was trained on the KITTI\textsubscript{v} dataset, and its flow estimation before refinement is shown in magenta (left). The optimized refined flow is presented in blue (center). We also \camrdy{show} the ground-truth scene flow in purple for reference (right). The refined flow better covers the target point cloud (top center ellipse). It also breaches the convex hull of the given target points and enables computing the correct flow for source points whose target is missing (bottom center ellipse).}
\label{fig:refine}
\end{figure}

\medskip
\noindent \textbf{The influence of flow refinement.} \quad Our flow results before and after refinement are presented in Figure~\ref{fig:refine}. Posing the learning part of \CorrFlow as a correspondence problem enables its effective training on a small dataset. However, the flow predictions from the training phase are confined to a linear combination of existing target points, which may not represent the exact flow of the scene. Moreover, wrong matches for the source points can occur and cause flow errors. In such cases, our refinement module comes into play.

Given the output flow from the trained correspondence model, the refinement module optimizes correction vectors subject to two objectives: a warped source point should be close to a target point; neighboring source points should have a similar flow. These objectives help fixing inconsistencies in the flow field and increase the flow accuracy. As seen in Figure~\ref{fig:refine}, our refinement step improves the initial flow estimation and results in an accurate flow field, which is similar to the ground-truth scene flow.

\subsection{Performance and Time Analysis} \label{subsec:perf_time_anal}
\begin{figure}[t!]
\centering
\includegraphics[width=\linewidth]{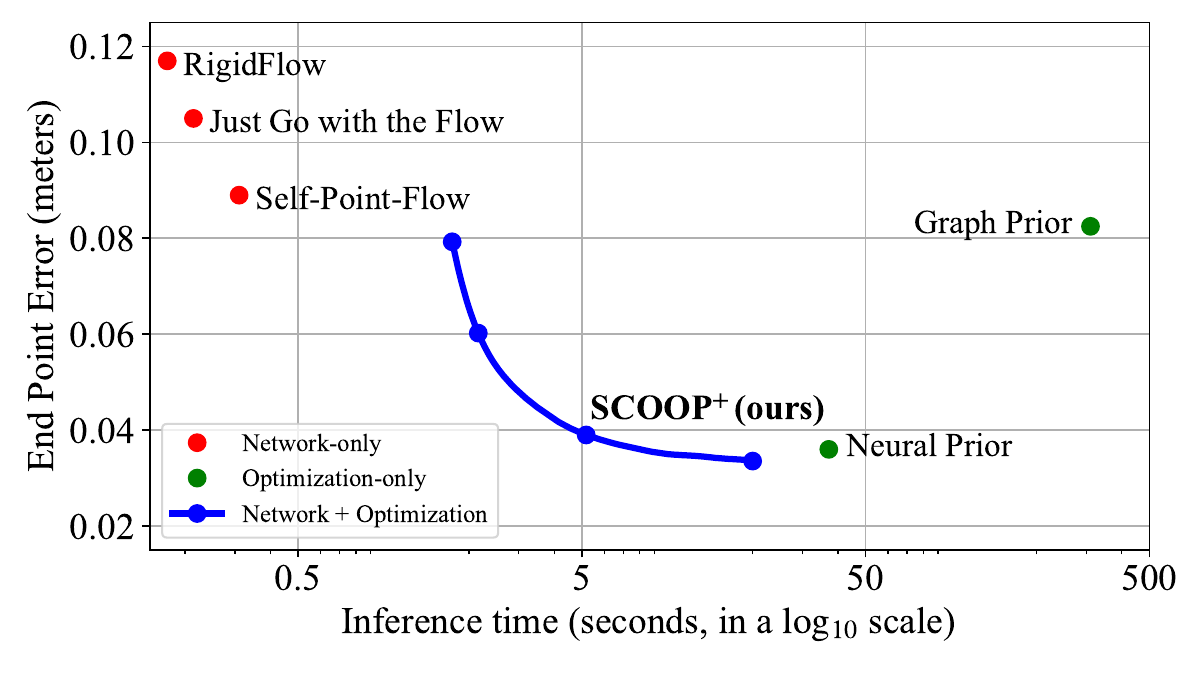}
\vspace{-20pt}
\caption{\textbf{Flow estimation error \vs inference time for the KITTI\textsubscript{t} dataset.} \CorrFlowwo\textsuperscript{+} has a lower error than the network-only models and a shorter inference time than the optimization-only methods. It also allows different balances along the error and time trade-off, as presented
by the blue curve.}
\vspace{5pt}
\label{fig:epe_vs_time}
\end{figure}

We analyze the performance-time trade-off in Figure~\ref{fig:epe_vs_time} by recording the $EPE$ and inference time for different methods. The measurements were done on an Nvidia Titan Xp GPU for computing the flow for \camrdy{complete point clouds of} the KITTI\textsubscript{t} dataset.

Network-only methods~\cite{mittal2020just, li2021self, li2022rigidflow} tend to be fast but with limited accuracy. Optimizing the flow prediction separately for each scene~\cite{li2021neural} results in a low $EPE$. However, it takes a long time. Our hybrid method bridges the trade-off gap between these two approaches. \camrdy{\CorrFlowwo\textsuperscript{+}} offers a working point with more than 50\% error reduction over the feed-forward models and \camrdy{about $8\times$} faster inference time than the optimization-only Neural Prior work. \camrdy{\CorrFlowwo\textsuperscript{+}} also enables a different balance between time and performance, as seen in Figure~\ref{fig:epe_vs_time}. By reducing the number of run-time optimization steps, the user can shorten the inference time, achieving a working point closer to that of the network-only models.

\subsection{Ablation Study} \label{subsec:ablation_study_scoop}

The design choices in our method are verified by ablation experiments presented in Table~\ref{tbl:ablation_method_elements}. We change one element each time and keep all the others the same. The following ablative settings were examined: (a) use all target points for soft correspondence instead of the ones with \camrdy{the} highest transport amount (Equation~\ref{eq:soft_corr}); (b) ignore the point matching confidence by setting $p_{x_i} = 1$ in Equation~\ref{eq:dist_loss} and $\alpha_{conf} = 0$ in Equation~\ref{eq:total_loss_scoop}; (c) exclude the smoothness flow loss $\Lcal_{flow}$ from Equation~\ref{eq:total_loss_scoop}; and (d) turn off the flow refinement module.

The ablation study validates the contribution of the proposed components to the method's performance. Considering a subset of target points for correspondence enables the model to concentrate on the most relevant candidates for flow estimation. The matching confidence emphasizes the influence of the more confident points in our confidence-aware distance loss $\Lcal_{dist}$. The smoothness loss term is important for regularizing the point representation learning to obtain similar features across neighboring points. Lastly, our flow refinement optimization improves the consistency of the flow field and reduces the $EPE$ substantially. \camrdy{In the supplementary material, we provide an ablation study on the FT3D\textsubscript{o} train set size and find that a 10\% fraction of the data suffices for our method to realize its potential.}

\begin{table}[tb!]
\small
\centering
\begin{tabular}{ @{ } l c c @{ } }
\toprule
Setting & $EPE\!\downarrow$ & $AS\!\uparrow$ \\
\midrule
(a) All target points as candidates ($k_s = n$)         & 0.047 & 91.1 \\
(b) W/O confidence ($p_{x_i} = 1$, $\alpha_{conf} = 0$) & 0.044 & 90.3 \\
(c) W/O smoothness loss term ($\alpha_{flow} = 0$)      & 0.056 & 86.6 \\
(d) W/O flow refinement ($R^* = 0$)                     & 0.115 & 43.8 \\
Our complete method                                     & \textbf{0.039} & \textbf{93.6} \\
\bottomrule
\end{tabular}
\caption{\textbf{Component ablative settings.} \CorrFlow was trained on KITTI\textsubscript{v} and evaluated on KITTI\textsubscript{t}. The results show that the best performance is obtained with our complete method. Additional details about the ablation experiments are given in subsection~\ref{subsec:ablation_study_scoop}.}
\label{tbl:ablation_method_elements}
\end{table}

\section{Conclusions} \label{sec:conclusions}
This paper presented \CorrFlowwo, a novel self-supervised scene flow estimation method for 3D point clouds based on correspondence learning and flow refinement optimization. Previous works suggested learning a flow regression model, training a neural network that jointly learned point cloud correspondence and flow refinement, or optimizing the flow completely at run-time without learning.

In contrast, we split the flow prediction process into two simpler problems. Our correspondence model is focused only on learning point features to initialize the flow from soft matches between the point clouds. Then, we directly optimize a residual flow refinement at run-time. This approach enables \CorrFlow to be trained on a small set of point cloud scenes without utilizing ground-truth supervision while outperforming state-of-the-art fully-supervised and self-supervised learning methods, as well as optimization-based alternative techniques.

\ifarxiv
{\small
\bibliographystyle{ieee_fullname.bst}
\bibliography{references.bib}
}

\clearpage
\appendix
\section*{Supplementary Material}
We provide more information regarding our flow estimation method \CorrFlowwo. Section~\ref{sec:optimal_transport_supp} presents the derivation of point cloud correspondence as an optimal transport problem and the solution by the Sinkhorn algorithm. Section~\ref{sec:results_supp} includes additional results for the experiments presented in the paper. In Section~\ref{sec:experiment_supp}, we report the results of an additional experiment on a non-occluded data version. Finally, section~\ref{sec:implementation_supp} elaborates on our implementation details, including network architecture, training and inference procedure, and the optimization settings of \CorrFlowwo.

\section{Correspondence as Optimal Transport} \label{sec:optimal_transport_supp}
As mentioned in the paper, our correspondence-based flow between the point clouds $X, Y \in \mathbb{R}^{n \times 3}$ builds on the optimal transport formulation presented in FLOT~\cite{puy2020flot}. For completeness, we briefly review the optimal transport problem and the Sinkhorn algorithm for solving it.

We begin with a hypothetical perfect case, where each source point $x_i \in X$ has an exact matching target point $y_j \in Y$. Thus, the flow field holds:
\begin{equation} \label{eq:perfect_flow}
X + F^* = \Pi Y,
\end{equation}

\noindent where $\Pi \in \{0, 1\}^{n \times n}$ is a permutation matrix representing the correspondence between the point clouds, with $\Pi_{ij} = 1$ if $x_i$ matches $y_j$ and $\Pi_{ij} = 0$ otherwise.

In this case, estimating the point correspondences can be modeled as an optimal transport problem~\cite{puy2020flot}. Assuming that each point in $X$ has a mass $\frac{1}{n}$ and each point in $Y$ receives a mass $\frac{1}{n}$, the optimal mass transport is given by:
\begin{equation} \label{eq:perfect_transport}
\begin{split}
T^* = &\argmin_{T \in \mathbb{R}_{+}^{n \times n}} \sum_{ij} C_{ij} T_{ij} \\
&\text{such that} \quad T 1_n = \frac{1}{n} 1_n, \quad T^\top 1_n = \frac{1}{n} 1_n,
\end{split}
\end{equation}

\noindent where $1_n \in \mathbb{R}^n$ is a vector with all entries equal 1, $C_{ij} \ge 0$ is the transport cost from point $x_i$ to point $y_j$, and $T_{ij} \ge 0$ is the amount of mass transported between these points. The two terms on the second row of Equation~\ref{eq:perfect_transport} are mass constraints, demanding that the total mass delivered from each source point and received by each target point is exactly $\frac{1}{n}$. $T^*$ is optimal \camrdy{in the sense that} the mass is transported from $X$ to $Y$ with minimal cost.

In practice, usually, there is no perfect match between the point clouds due to objects appearing in or disappearing from the scene or different points sampled on the scene's surface, and the mass constraints in Equation~\ref{eq:perfect_transport} do not hold. Thus, instead of Equation~\ref{eq:perfect_transport}, we used the relaxed version of the transport problem presented in Equation~\ref{eq:regularized_transport} in the paper. The relaxed transport problem is solved by the Sinkhorn algorithm~\cite{cuturi2013sinkhorn, chizat2018scaling}, which estimates $T^*$ from $C$. We provide the algorithm's details in Algorithm~\ref{alg:sinkhorn}. In our implementation, the number of iterations $M$ is set to 1.

\begin{figure}[t!]
\centering
\includegraphics[width=\linewidth]{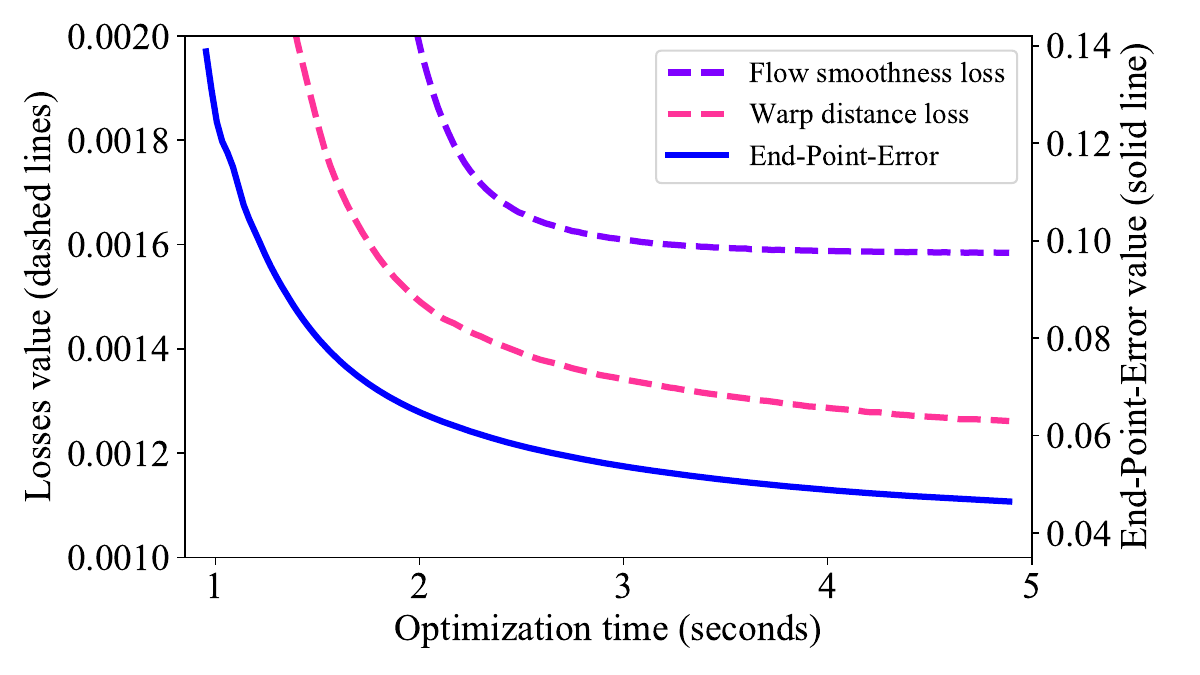}
\vspace{-20pt}
\caption{\textbf{Refinement evolution.} \CorrFlow was trained on FT3D\textsubscript{o} and evaluated on KITTI\textsubscript{o}. We present the refinement loss values and the corresponding End-Point-Error ($EPE$) during the optimization process. The losses are effectively minimized and result in a substantial reduction of the flow estimation error.}
\vspace{10pt}
\label{fig:refinement_evolution_supp}
\end{figure}

\begin{table}[t!]
\small
\centering
\begin{tabular}{l c c @{ } c @{ } c @{ } c @{ }}
\toprule
Method & Refinement & $\:EPE\!\downarrow$ & $\:AS\!\uparrow$ & $\:AR\!\uparrow$ & $\:Out.\!\downarrow$ \\
\midrule
FLOT~\cite{puy2020flot} & $\xmark$ & 0.142 & 30.6 & 61.9 & 57.6 \\
FLOT~\cite{puy2020flot} & $\cmark$ & \textbf{0.048} & \textbf{89.0} & \textbf{93.5} & \textbf{20.4} \\
\midrule
\CorrFlowwo\textsuperscript{+} (ours) & $\xmark$ & 0.139 & 36.1 & 63.6 & 54.9 \\
\CorrFlowwo\textsuperscript{+} (ours) & $\cmark$ & \textbf{0.047} & \textbf{91.3} & \textbf{95.0} & \textbf{18.6} \\
\bottomrule
\end{tabular}
\caption{\textbf{Our refinement optimization for another method.} FLOT and \CorrFlow were trained on 1,800 examples from FT3D\textsubscript{o} and tested on KITTI\textsubscript{o}, without or with our flow refinement component. The proposed refinement module considerably improves the flow estimation performance for both methods.}
\label{tbl:refinement_supp}
\end{table}

\section{Additional Results} \label{sec:results_supp}

\subsection{Refinement Evolution} \label{subsec:refinement_evo_supp}
We examine the relationship between our self-supervised losses in the flow refinement process, given in Equation~\ref{eq:flow_refinement}, and the resulting End-Point-Error metric ($EPE$), defined in subsection~\ref{subsec:experimental_setup}. Figure~\ref{fig:refinement_evolution_supp} shows the results (for better visualization, we multiply the smoothness loss value by a factor of $4 \cdot 10^{-2}$). During run-time, we minimize our smoothness and distance losses without using ground-truth flow labels. As a byproduct, the $EPE$ is reduced as well. This experiment implies that our refinement objective in Equation~\ref{eq:flow_refinement} correlates with the flow estimation error and serves as a good proxy for its minimization.

\subsection{Refinement Optimization for Another Method} \label{subsec:refine_another_supp}
A natural question is whether a flow estimation method other than ours can benefit from \camrdy{the proposed} refinement optimization module. To address this question, we trained FLOT~\cite{puy2020flot} on 1,800 examples from the FT3D\textsubscript{o} train set, as done for our method. Then, we evaluated FLOT's performance on the KITTI\textsubscript{o} data without or with our run-time refinement (with correspondence confidence equal to 1 for all the source points). Table~\ref{tbl:refinement_supp} summarizes the results.

Training on a 10\% fraction of FT3D\textsubscript{o} \camrdy{data} degrades FLOT's performance in comparison to using the complete dataset, as reported in Table~\ref{tbl:kitti} in the main body. However, our refinement optimization substantially contributes to the flow precision of FLOT and even yields better results compared to using the whole training set. This experiment hints that our proposed run-time refinement is not tailor-made for \CorrFlow and can benefit another method as well.

\subsection{Qualitative Results} \label{subsec:qualitative_supp}
In Figure~\ref{fig:additional_supp}, we present additional results of \CorrFlow for KITTI\textsubscript{o} data for various challenging cases. For example, our method can gracefully handle different point densities, as cars with varying distances from the LiDAR sensor exhibit. In addition, since we require consistency of the flow field over the point cloud, \CorrFlow can correctly estimate the flow for an object with a repetitive structure, such as a fence. At the same time, our flow estimation method is versatile. It copes with shapes of different geometry and size, such as the pole and the facade. \CorrFlow can also predict translation vectors of different directions and magnitudes, as for the car and pole.

\RestyleAlgo{ruled}
\begin{algorithm}[tb!]
\KwData{$\text{cost matrix}\; C,\: \text{parameters}\; \epsilon, \lambda \ge 0,\: M > 0.$}
\KwResult{$\text{optimal transport matrix}\; T^*.$}
$T \gets \exp (-C/\epsilon)$\;
$a \gets \frac{1}{n} 1_n$\;
\For{$m = 1, \dots, M$}
{
$b \gets (\frac{1}{n} 1_n / (T^\top \! a))^{\lambda/(\lambda + \epsilon)}$\;
$a \gets (\frac{1}{n} 1_n / (T b))^{\lambda/(\lambda + \epsilon)}$\;
}
$T^* \gets \diag(a)\:T\diag(b)$\;
\caption{\textbf{The Sinkhorn Algorithm.}} \label{alg:sinkhorn}
\end{algorithm}

\subsection{Ablation Runs} \label{subsec:ablation_supp}
\camrdy{In Table~\ref{tbl:ablation_train_size}, we report results of our method for different train set sizes of FT3D\textsubscript{o}. The table shows that a 10\% fraction of the FT3D\textsubscript{o} data is sufficient for \CorrFlow to converge to its optimal performance.}

Table~\ref{tbl:ablation_supp} presents additional ablation experiments. In this round, we examined the following settings (one configuration change at a time). (a) Turn off the Sinkhorn normalization. In this case, we used $T = \exp (-C/\epsilon)$ instead of $T^*$ from Algorithm~\ref{alg:sinkhorn}, and the correspondence construction in Equations~\ref{eq:corr_weights} and \ref{eq:soft_corr} was done with target points with minimal matching cost $C$ rather than maximal transport $T^*$. (b) Apply a higher number of iterations in the Sinkhorn algorithm by setting $M = 3$ instead of $M = 1$. (c) Linear normalization for the correspondence confidence $p_{x_i} = (s_{x_i} + 1)/2$ instead of the non-linear truncation $p_{x_i} = \max(s_{x_i}, 0)$. In all these settings, the difference in the method's performance was small, implying its robustness to such configuration changes.

\begin{table}[tb!]
\small
\centering
\begin{tabular}{ @{ } l | c c c @{ } }
\toprule
FT3D\textsubscript{o} number of training examples & 180 & 1,800 & 18,000 \\
\midrule
KITTI\textsubscript{o} $EPE\!\downarrow$ & 0.057 & 0.047 & 0.047 \\
\bottomrule
\end{tabular}
\caption{\textbf{Train set size ablation.} We trained \CorrFlow on the FT3D\textsubscript{o} dataset using a different number of instances and measured the $EPE$ on the KITTI\textsubscript{o} dataset. A subset of only 1,800 training examples \camrdy{is sufficient for our technique.}}
\label{tbl:ablation_train_size}
\end{table}

\begin{table}[t!]
\small
\centering
\begin{tabular}{l c @{ } c @{ } c @{ } c @{ }}
\toprule
Setting & $\:EPE\!\downarrow$ & $\:AS\!\uparrow$ & $\:AR\!\uparrow$ & $\:Out.\!\downarrow$ \\
\midrule
(a) W/O Sinkhorn & 0.042 & 91.6 & 95.9 & 16.1 \\
(b) 3 Sinkhorn iterations & 0.040 & 92.9 & 96.4 & 15.3 \\
(c) Linear $p_{x_i}$ normalization & 0.040 & 93.5 & 96.4 & 15.5 \\
The proposed method & \textbf{0.039} & \textbf{93.6} & \textbf{96.5} & \textbf{15.2} \\
\bottomrule
\end{tabular}
\vspace{0.1cm}
\caption{\textbf{Additional ablations.} We trained \CorrFlow with different configurations on KITTI\textsubscript{v} and evaluated its performance on KITTI\textsubscript{t}. The table shows that our method is robust to these configuration variations. Details about the ablative settings appear in subsection~\ref{subsec:ablation_supp}.}
\label{tbl:ablation_supp}
\end{table}

\subsection{Limitation} \label{subsec:limitation_supp}
A failure case of \CorrFlow is presented in Figure~\ref{fig:failure_supp}. When a part of the source scene is completely missing from the target, the correspondence to existing target points is inaccurate, and the flow predicted by our method does not represent the motion of that part. In future work, we plan to detect such wrong matches by remaining inconsistencies in the flow field and leverage the global motion of the scene to deduce the flow for completely occluded regions.

\begin{figure*}[t!]
\centering
\includegraphics[width=\linewidth]{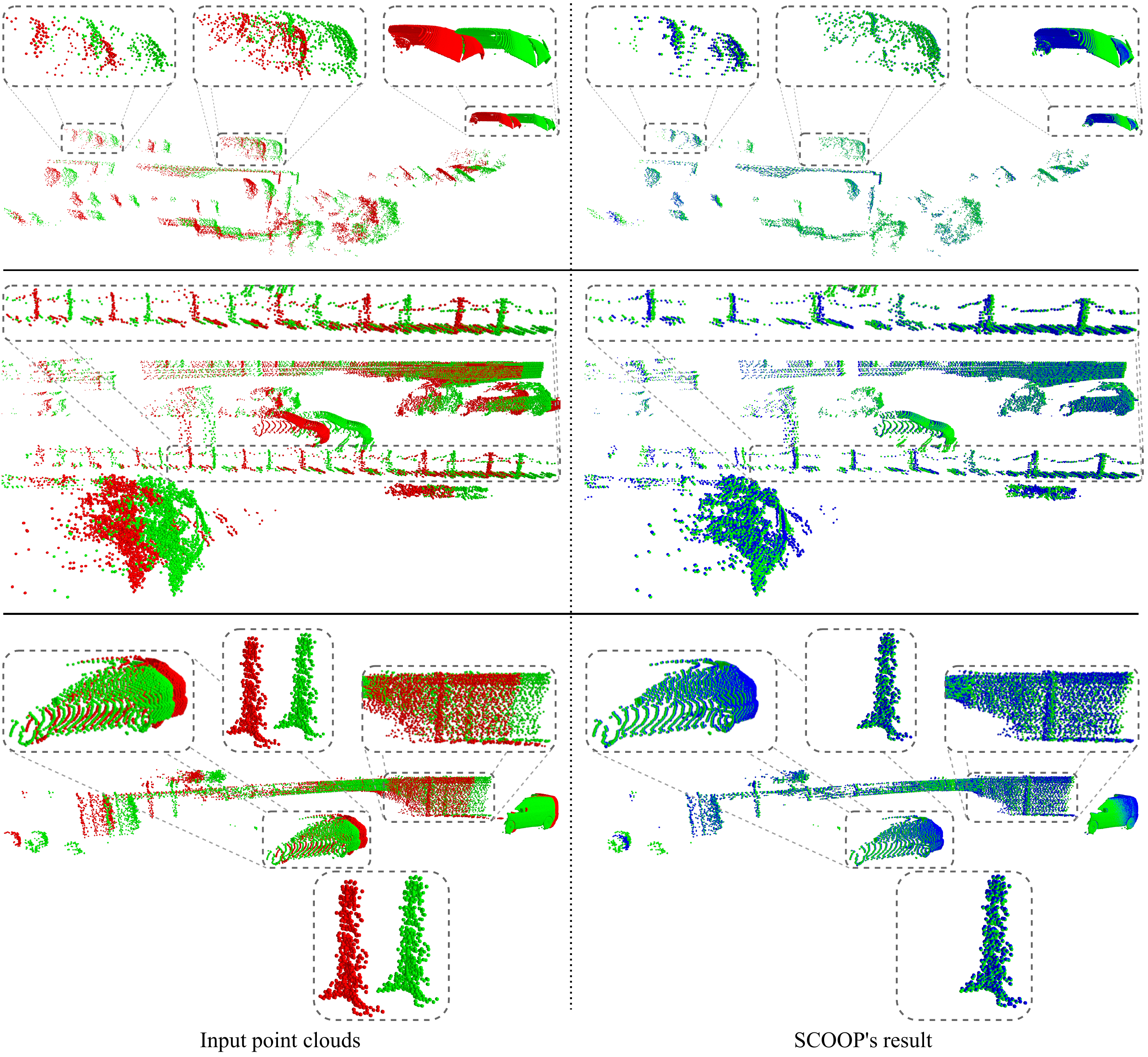}
\caption{\textbf{Visual results.} We applied \CorrFlow to different LiDAR scenes. The source and target input point clouds are presented in red and green, respectively, and the warped source is shown in blue. Our method is able to \camrdy{predict} the scene flow in a variety of challenging scenarios, such as varied point cloud density (top), repetitive structures (middle), and objects with different sizes and motions (bottom).}
\label{fig:additional_supp}
\end{figure*}

\section{An Additional Experiment} \label{sec:experiment_supp}
\camrdy{In addition to FT3D\textsubscript{o} and KITTI\textsubscript{o}, Gu \etal~\cite{gu2019hplflownet} prepared another point cloud version of the FlyingThings3D and KITTI datasets, denoted as FT3D\textsubscript{s} and KITTI\textsubscript{s}, respectively. In their version, all occluded points are removed, and each source point has a matched target point. This version of the datasets is also popular in the scene flow literature, and for a comprehensive evaluation, we report our method's results for this case as well. Additional details about the datasets appear in subsection~\ref{subsec:training_supp}.}

Since the \camrdy{point clouds produced by Gu \etal have} no occlusions, we adapt our method to the nature of this data. Instead of the distance loss from Equation~\ref{eq:dist_loss}, we use the bidirectional Chamfer Distance loss~\cite{wu2020pointpwcnet, lang2021dpc}:
\begin{equation} \label{eq:dist_loss_supp}
\begin{split}
\Lcal_{cd}& = CD(\widehat{Y}, Y) = \\
  &\frac{1}{|\widehat{Y}|}\sum_{\hat{y} \in \widehat{Y}}{\min_{y \in Y}||\hat{y} - y||_2^2} +
   \frac{1}{|Y|}\sum_{y \in Y}{\min_{\hat{y} \in \widehat{Y}}||y - \hat{y}||_2^2},
\end{split}
\end{equation}

\noindent where $\widehat{Y}$ is the softy corresponding point cloud to the source point cloud $X$ (from Equation~\ref{eq:flow_field}), and $Y$ is the target point cloud. The Chamfer Distance is also used in the refinement process and replaces the first term in the optimization objective in Equation~\ref{eq:flow_refinement}. If we define $\widehat{Y}_r = \widehat{Y} + R$, the updated distance loss term for the flow refinement optimization is $CD(\widehat{Y}_r, Y)$. The rest of our method's formulation remains the same.

Following the evaluation protocol of previous work~\cite{gu2019hplflownet, puy2020flot, kittenplon2021flowstep3d}, we train \CorrFlow on FT3D\textsubscript{s} and evaluate the performance on the test set of FT3D\textsubscript{s} and on the KITTI\textsubscript{s} data. Different from prior work, we use \camrdy{only} 10\% of \camrdy{the} FT3D\textsubscript{s} training data, which suffices for our correspondence model to coverage. The evaluation metrics are the same as those in the main body, detailed in subsection~\ref{subsec:experimental_setup}. For both training and testing, we use point clouds with $n = 8192$ points.

Tables~\ref{tbl:ft3d_s_supp} and~\ref{tbl:kitti_s_supp} present our test results for FT3D\textsubscript{s} and KITTI\textsubscript{s}, respectively, compared to abundant recent alternative methods. While trained \camrdy{only on} a 10\% \camrdy{fraction} of the data, \CorrFlow achieves competitive results compared to other self-supervised methods on FT3D\textsubscript{s}. On the KITTI\textsubscript{s} dataset, we surpass the performance of both self and fully-supervised methods for all the evaluation metrics. For example, \CorrFlow improves the $EPE$ metric by 37\% over the very recent Bi-PointFlowNet work~\cite{cheng2022bi-pointflownet}, reducing the flow estimation error from $0.030$ to $0.019$ meters. These results suggest that our method is highly effective for the real-world KITTI\textsubscript{s} data.

\begin{table}[t!]
\small
\centering
\begin{tabular}{l c c @{ } c @{ } c @{ } c @{ }}
\toprule
Method & Sup. & $\:EPE\!\downarrow$ & $\:AS\!\uparrow$ & $\:AR\!\uparrow$ & $\:Out.\!\downarrow$ \\
\midrule
FlowNet3D~\cite{liu2019flownet3d} & \textit{Full} & 0.114 & 41.3 & 77.1 & 60.2 \\
HPLFlowNet~\cite{gu2019hplflownet} & \textit{Full} & 0.080 & 61.4 & 85.6 & 42.9 \\
PointPWC-Net~\cite{wu2020pointpwcnet} & \textit{Full} & 0.059 & 73.8 & 92.8 & 34.2 \\
FLOT~\cite{puy2020flot} & \textit{Full} & 0.052 & 73.2 & 92.7 & 35.7 \\
PV-RAFT~\cite{wei2021pv-raft} & \textit{Full} & 0.046 & 81.7 & 95.7 & 29.4 \\
FlowStep3D~\cite{kittenplon2021flowstep3d} & \textit{Full} & 0.046 & 81.6 & 96.1 & 21.7 \\
HCRF-Flow~\cite{li2021hcrf-flow} & \textit{Full} & 0.049 & 83.4 & 95.1 & 26.1 \\
RCP~\cite{gu2022rcp} & \textit{Full} & 0.040 & 85.7 & 96.4 & 19.8 \\
Rigid3DSceneFlow~\cite{gojcic2021weakly} & \textit{Full} & 0.052 & 74.6 & 93.6 & 36.1 \\
3D-OGFlow~\cite{ouyang2021occlusion} & \textit{Full} & 0.036 & 87.9 & - & 19.7 \\
SCTN~\cite{li2022sctn} & \textit{Full} & 0.038 & 84.7 & 96.8 & 26.8 \\
3DFlow~\cite{wang2022what} & \textit{Full} & \textbf{0.028} & \textbf{92.9} & \textbf{98.2} & 14.6 \\
Bi-PointFlowNet~\cite{cheng2022bi-pointflownet} & \textit{Full} & \textbf{0.028} & 91.8 & 97.8 & \textbf{14.3} \\
\midrule
Ego-motion~\cite{tishchenko2020self} & \textit{Self} & 0.170 & 25.3 & 55.0 & 80.5 \\
PointPWC-Net~\cite{wu2020pointpwcnet} & \textit{Self} & 0.121 & 32.4 &67.4 &68.8 \\
Self-Point-Flow~\cite{li2021self} & \textit{Self} & 0.101 & 42.3 & 77.5 & 60.6 \\
FlowStep3D~\cite{kittenplon2021flowstep3d} & \textit{Self} & 0.085 & 53.6 & 82.6 & 42.0 \\
RSFNet~\cite{he2022self} & \textit{Self} & 0.075 & 58.9 & 86.2 & 47.0 \\
RCP~\cite{gu2022rcp} & \textit{Self} & 0.077 & 58.6 & 86.0 & 41.4 \\
RigidFlow~\cite{li2022rigidflow} & \textit{Self} & 0.069 & 59.6 & 87.1 & 46.4 \\
\CorrFlow (ours) & \textit{Self} & 0.084 & 56.7 & 85.1 & 48.5 \\
\bottomrule
\end{tabular}
\caption{\textbf{Quantitative comparison on the FT3D\textsubscript{s} test set.} All the methods were trained on the train split of FT3D\textsubscript{s}. Our method is on par with other self-supervised methods.}
\label{tbl:ft3d_s_supp}
\end{table}

\begin{table}[t!]
\small
\centering
\begin{tabular}{l c c @{ } c @{ } c @{ } c @{ }}
\toprule
Method & Sup. & $\:EPE\!\downarrow$ & $\:AS\!\uparrow$ & $\:AR\!\uparrow$ & $\:Out.\!\downarrow$ \\
\midrule
FlowNet3D~\cite{liu2019flownet3d} & \textit{Full} & 0.177 & 37.4 & 66.8 & 52.7 \\
HPLFlowNet~\cite{gu2019hplflownet} & \textit{Full} & 0.117 & 47.8 & 77.8 & 41.0 \\
PointPWC-Net~\cite{wu2020pointpwcnet} & \textit{Full} & 0.069 & 72.8 & 88.8 & 26.5 \\
FLOT~\cite{puy2020flot} & \textit{Full} & 0.056 & 75.5 & 90.8 & 24.2 \\
PV-RAFT~\cite{wei2021pv-raft} & \textit{Full} & 0.056 & 82.3 & 93.7 & 21.6 \\
FlowStep3D~\cite{kittenplon2021flowstep3d} & \textit{Full} & 0.055 & 80.5 & 92.5 & 14.9 \\
HCRF-Flow~\cite{li2021hcrf-flow} & \textit{Full} & 0.053 & 86.3 & 94.4 & 18.0 \\
RCP~\cite{gu2022rcp} & \textit{Full} & 0.048 & 84.9 & 94.5 & 12.3 \\
Rigid3DSceneFlow~\cite{gojcic2021weakly} & \textit{Full} & 0.042 & 84.9 & 95.9 & 20.8 \\
3D-OGFlow~\cite{ouyang2021occlusion} & \textit{Full} & 0.039 & 88.2 & - & 17.5 \\
SCTN~\cite{li2022sctn} & \textit{Full} & 0.037 & 87.3 & 95.9 & 17.9 \\
3DFlow~\cite{wang2022what} & \textit{Full} & 0.031 & 90.5 & 95.8 & 16.1 \\
Bi-PointFlowNet~\cite{cheng2022bi-pointflownet} & \textit{Full} & 0.030 & 92.0 & 96.0 & 14.1 \\
\midrule
Ego-motion~\cite{tishchenko2020self} & \textit{Self} & 0.415 & 22.1 & 37.2 & 81.0 \\
PointPWC-Net~\cite{wu2020pointpwcnet} & \textit{Self} & 0.255 & 23.8 & 49.6 & 68.6 \\
Self-Point-Flow~\cite{li2021self} & \textit{Self} & 0.112 & 52.8 & 79.4 & 40.9 \\
FlowStep3D~\cite{kittenplon2021flowstep3d} & \textit{Self} & 0.102 & 70.8 & 83.9 & 24.6 \\
RSFNet~\cite{he2022self} & \textit{Self} & 0.092 & 74.7 & 87.0 & 28.3 \\
RCP~\cite{gu2022rcp} & \textit{Self} & 0.076 & 78.6 & 89.2 & 18.5 \\
RigidFlow~\cite{li2022rigidflow} & \textit{Self} & 0.062 & 72.4 & 89.2 & 26.2 \\
\CorrFlow (ours) & \textit{Self} & \textbf{0.019} & \textbf{97.1} & \textbf{98.5} & \textbf{10.7} \\
\bottomrule
\end{tabular}
\caption{\textbf{Quantitative comparison on the KITTI\textsubscript{s} data.} All the methods were trained on the train split of FT3D\textsubscript{s}. \CorrFlow outperforms all the compared alternatives, both the self-supervised and the fully-supervised ones.}
\label{tbl:kitti_s_supp}
\end{table}

\section{Implementation Details} \label{sec:implementation_supp}

\subsection{Network Architecture} \label{subsec:architecture_supp}
The point feature extraction is done by a neural network based on the PointNet++ architecture~\cite{qi2017pointnetpp}. The network includes 3 set-convolution layers, which increase the feature channels per point. Each layer contains a multi-layer perceptron, interleaved with instance normalization and a leaky ReLU activation with a negative slope of $-0.1$. After each convolutional layer, the point features are aggregated by a max pooling operation from 32 Euclidean nearest neighbor points. The coordinate difference between the point and its neighbors is concatenated to the input features of every set-convolution layer. Table~\ref{tbl:architecture_supp} details the feature dimensions of the network's layers.

\begin{figure}[t!]
\centering
\includegraphics[width=\linewidth]{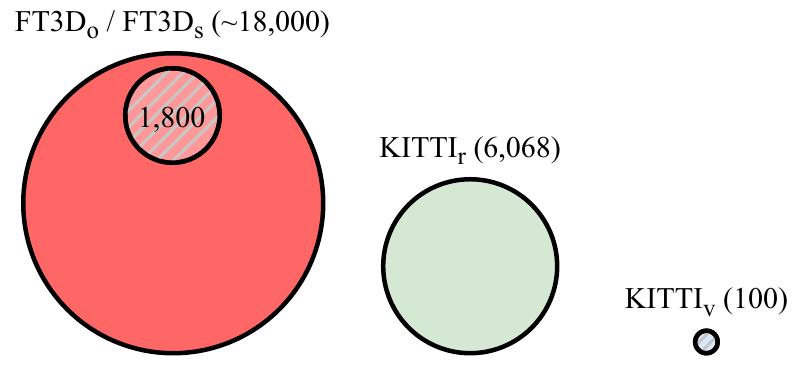}
\caption{\textbf{Visual illustration of the training datasets' size.} We use a small amount of data for training (stripe pattern) compared to the amount used by others (solid pattern).}
\label{fig:datasets_size_supp}
\end{figure}

\begin{figure*}[t!]
\centering
\includegraphics[width=\linewidth]{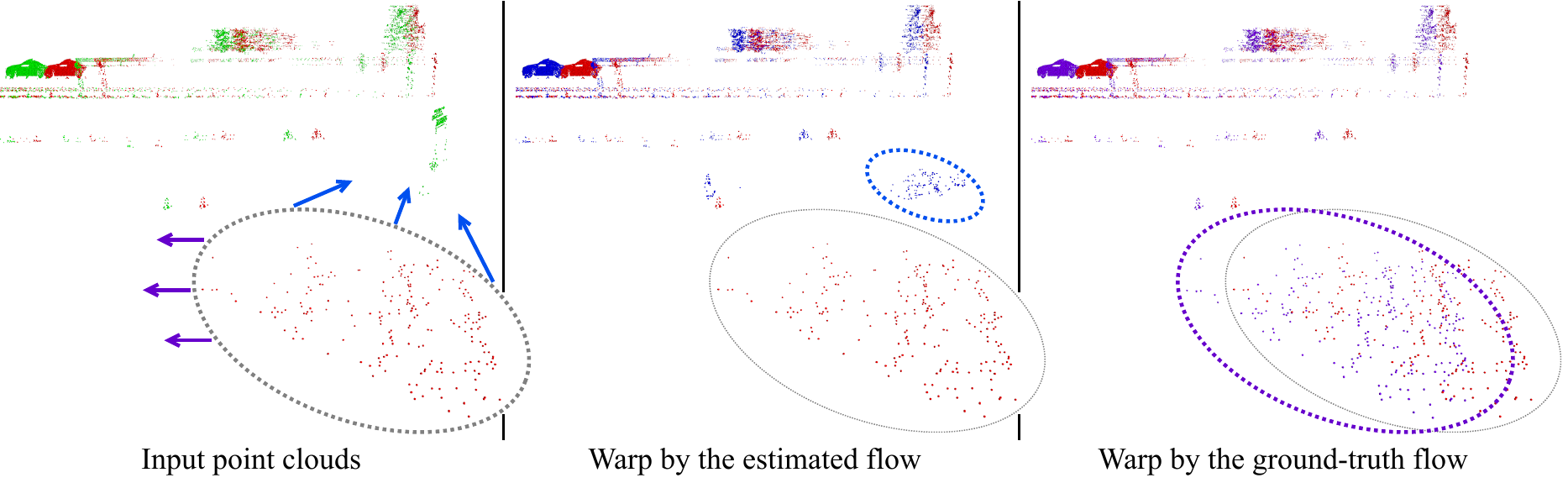}
\caption{\textbf{A failure example.} We show the source point cloud in red, the target in green (left), the translated source by \CorrFlow in blue (middle), and the translated source by the ground-truth flow in purple (right). A set of source points whose target is completely occluded is marked with a gray ellipse. Its warp by the estimated and the ground-truth flow is delineated by a blue ellipse and a purple ellipse, respectively. Our method struggles to predict the correct flow in such a case.}
\label{fig:failure_supp}
\end{figure*}

\subsection{Training and Inference} \label{subsec:training_supp}

\noindent \textbf{Training dataset size.} \quad We illustrate the size of the training datasets in Figure~\ref{fig:datasets_size_supp}. As explained in the paper (subsection~\ref{subsec:flow_res}), \CorrFlow is a data-light method that requires much less training data than other learning-based methods.

\medskip
\noindent \textbf{Occluded data version.} \quad
The FT3D\textsubscript{o} dataset contains point clouds of 8,192 points, where the z-axis coincides with the depth axis, and the maximal z-value is limited to 35 meters~\cite{liu2019flownet3d}. In the KITTI\textsubscript{o} dataset, there are several tens of thousands of points per \camrdy{scene}, with a different number of points for the source and target point clouds, denoted as $N_s$ and $N_t$, respectively. We align the z-axis of KITTI\textsubscript{o} to the depth axis and trim the maximal z-value to 35 meters, as done for the FT3D\textsubscript{o} data~\cite{puy2020flot}.

\begin{table}[tb!]
\small
\centering
\begin{tabular}{ @{ } c @{ } }
\toprule
Network architecture \\
\midrule
$concat(\text{coordinates}\:(3), \text{neighbors' coordinate difference}\:(3))$ \\
$SetConv(32, 32, 32)$ \\
$neighbors\;max\;pooling\:(32)$ \\
$concat(\text{features}\:(32), \text{neighbors' coordinate difference}\:(3))$ \\
$SetConv(64, 64, 64)$ \\
$neighbors\;max\;pooling\:(64)$ \\
$concat(\text{features}\:(64), \text{neighbors' coordinate difference}\:(3))$ \\
$SetConv(128, 128, 128)$ \\
$neighbors\;max\;pooling\:(128)$ \\
\bottomrule
\end{tabular}
\vspace{0.1cm}
\caption{\textbf{The architecture of the feature extraction model.} The values in parentheses indicate the per-point feature dimension at each network stage. $concat$ represents a concatenation operation. The coordinate difference and max pooling operation are computed with a neighborhood of 32 nearest points in the Euclidean space. $SetConv$ is the set convolution described in subsection~\ref{subsec:architecture_supp}, where the numbers in its parentheses refer to the filter sizes of the multi-layer perceptron.}
\label{tbl:architecture_supp}
\end{table}

For memory-efficient training, \CorrFlow is trained on \camrdy{point} sets with the same number of $n = 2,\!048$ points sampled at random from the original point clouds. \camrdy{Following previous work~\cite{liu2019flownet3d, puy2020flot, mittal2020just, li2022rigidflow}, we evaluate \CorrFlow on small test point clouds of randomly sampled 2,048 points.} However, we also employ our method to infer the flow for all the points $N_s$ in the source point cloud, as explained next.

At the test-time, we randomly shuffle the source points and the target points, divide them into disjoint chunks of $n = 2,\!048$ points, and compute the point features $\Phi_X$ and $\Phi_Y$ \camrdy{for each chunk}, as done in the training stage. If the number of points is not divided by $n$, we pad with randomly selected points from within the point cloud to the closest multiple of $n$. Then, for each source chuck, we \camrdy{calculate} the matching cost with respect to all the points in the target, obtain a cost matrix $C_{chunk} \in \mathbb{R}^{n \times N_t}$, and compute \camrdy{the} correspondence-based flow $F_{chunk} \in \mathbb{R}^{n \times 3}$. Afterward, we collect the flow from the different chunks, remove the padded points (if any), and get the per-point flow $F \in \mathbb{R}^{N_s \times 3}$.

\begin{table}[t!]
\small
\centering
\begin{tabular}{l c c c c}
\toprule
\multirow{2}{*}{Train/Test data (\#points)} & \multirow{2}{*}{$k_f$} & \multirow{2}{*}{$\lambda_{flow}$} & Gradient & Update \\
 & & & steps & rate \\
\midrule
\camrdy{FT3D\textsubscript{o}/KITTI\textsubscript{o} (2,048)} & \camrdy{32} & \camrdy{1.0} & \camrdy{1000} & \camrdy{0.05} \\
\camrdy{KITTI\textsubscript{v}/KITTI\textsubscript{t} (2,048)} & \camrdy{32} & \camrdy{1.0} & \camrdy{1000} & \camrdy{0.05} \\
FT3D\textsubscript{o}/KITTI\textsubscript{o} (29,951) & 32 & 1.0 & 150 & 0.2 \\
KITTI\textsubscript{v}/KITTI\textsubscript{t} (30,814) & 32 & 1.0 & 150 & 0.2 \\
FT3D\textsubscript{s}/FT3D\textsubscript{s} (8,192) & 16 & 1.0 & 1000 & 0.1 \\
FT3D\textsubscript{s}/KITTI\textsubscript{s} (8,192) & 32 & 1.0 & 1000 & 0.05 \\
\bottomrule
\end{tabular}
\vspace{0.1cm}
\caption{\textbf{Refinement hyperparameters.} The table details the values we used for our flow refinement optimization process for different dataset settings. \camrdy{For each setting, we indicate the train/test datasets and the average number of points in the test point clouds.}}
\label{tbl:refinement_hyperparams_supp}
\end{table}

Our inference process \camrdy{for the complete point clouds} has several advantages. First, it can be used for source and target point clouds with different \camrdy{cardinality} since each point cloud is padded to a multiple of $n$. Second, as we extract point features in chunks of $n$ points, the process remains memory-efficient and emulates inputs to the network similar to the training phase. Third, it utilizes the complete point information from the target by computing the cost matrix and correspondence flow \camrdy{at} the original target point cloud resolution.

Similarly, we perform the flow refinement optimization \camrdy{at} the full source and target point cloud resolution. Namely, the distance loss for flow refinement is computed between the complete warped source and the complete target, and the flow smoothness loss is calculated at the original source point cloud resolution. \camrdy{This way, the whole scene data is exploited.}

\camrdy{For network-only baselines~\cite{mittal2020just, li2021self, li2022rigidflow}, inferring the scene flow directly for the high point cloud resolution is computationally infeasible, let alone training the models on the complete large point clouds. Thus, following their training scheme on small point clouds with 2,048 points, we divided the original point clouds into chunks of 2,048 points, applied the models, and averaged the results across the chunks to obtain the evaluation for all the points in the dataset.}

\camrdy{We note that our results for Neural Prior~\cite{li2021neural} are different from those reported in their paper. In their work, they did not limit the depth value of the point clouds. However, in our work, we used points with a maximal depth of 35 meters to align with previous learning-based methods~\cite{liu2019flownet3d, puy2020flot}.}

\medskip
\noindent \textbf{Non-occluded data version.} \quad
The FT3D\textsubscript{s} dataset has 19,640 and 3,824 point cloud pairs for the train and test sets, respectively. Each point cloud has 8,192 points. We keep aside 2,000 examples from the training set for validation during training. The KITTI\textsubscript{s} data include 200 pairs of source and target point clouds, where 142 of which are used for evaluation. Ground points are removed by a threshold on the height. In both datasets, points with a depth larger than 35 meters are excluded, as done by Gu \etal~\cite{gu2019hplflownet}. For testing, we randomly sample 8,192 points from the source and target point clouds each. Our inference time for FT3D\textsubscript{s} and KITTI\textsubscript{s} is about 3.7 seconds.

\subsection{Optimization} \label{subsec:optimization_supp}
We trained \camrdy{our} point embedding model with an ADAM optimizer with an initial learning rate of 0.001 and a momentum of 0.9. On the FT3D\textsubscript{o} dataset, we trained the model for 30, 100, and 200 epochs when using 180, 1,800, and 18,000 training examples, respectively. For training on the KITTI\textsubscript{v} dataset, we used 400 epochs, and the learning rate was reduced by a factor of 10 after 340 epochs. In all these cases, the batch size was 4. For the FT3D\textsubscript{s} dataset, we selected 1,800 examples at random and trained our model for 60 epochs with a batch size of 1. The learning rate was multiplied by 0.1 after 50 epochs. 

As mentioned in the paper, we optimized $\epsilon$ and $\lambda$ from the regularized transport problem (Equation~\ref{eq:regularized_transport} in the main body) during the training process. Their $\log$ value was learned to ensure their non-negativity. In addition, we added a constant of 0.03 to the learned value of $\epsilon$ for the numerical stability of the learning process.

The refinement component $R^*$ in Equation~\ref{eq:flow_refinement} in the paper was defined as an optimizable variable and initialized to a matrix of zeros. We optimized its value using an ADAM optimizer with a momentum of 0.9. Further hyperparameters are given in Table~\ref{tbl:refinement_hyperparams_supp}. All our experiments were done on an NVIDIA Titan Xp GPU.

\else
\ifmainpaper
{\small
\bibliographystyle{ieee_fullname.bst}
\bibliography{references.bib}
}

\else
\newpage
\clearpage
dummy
\clearpage
dummy

{\small
\bibliographystyle{ieee_fullname.bst}
\bibliography{references_supp.bib}
}
\fi
\fi

\end{document}